\pdfoutput=1

\documentclass[11pt]{article}

\usepackage[]{acl} %

\usepackage{times}
\usepackage{latexsym}

\usepackage{multirow}
\usepackage{booktabs}
\usepackage{graphicx}
\usepackage{inconsolata}

\usepackage[T1]{fontenc}

\usepackage[utf8]{inputenc}

\usepackage{microtype}

\usepackage{array}
\usepackage{listings}
\usepackage[normalem]{ulem}

\newsavebox\myv

\definecolor{backcolour}{RGB}{245,248,250}
\definecolor{emph}{RGB}{166,88,53}
\definecolor{nightblue}{RGB}{9,49,105}
\definecolor{keywords}{RGB}{207,33,46}
\definecolor{lightpurple}{RGB}{130,81,223}
\definecolor{codegreen}{RGB}{1,113,1}
\definecolor{complexred}{RGB}{180,23,0}
\definecolor{stepgreen}{RGB}{1,113,1}

\lstdefinestyle{prompt}{
    backgroundcolor=\color{backcolour},
    commentstyle=\color{codegreen},
    keywordstyle=\color{keywords},
    stringstyle=\color{nightblue},
    basicstyle=\fontsize{7}{8}\ttfamily,
    breakatwhitespace=true,
    breaklines=true,
    captionpos=b,
    keepspaces=true,
    numberstyle=\tiny\color{codegray},
    numbersep=2pt,
    showspaces=false,
    showstringspaces=false,
    showtabs=false,
    tabsize=1,
    linewidth=0.98\columnwidth,
    frame=tb,
    xrightmargin=0pt,
    xleftmargin=0.23cm,
    numbers=none,
    aboveskip=0.4cm,
    belowskip=0.4cm,
    keywordstyle=\color{blue},
    language=Scala,
    morekeywords={extension,using},
}

\lstdefinestyle{promptexample}{
    backgroundcolor=\color{backcolour},
    commentstyle=\color{codegreen},
    keywordstyle=\color{keywords},
    stringstyle=\color{nightblue},
    basicstyle=\fontsize{7}{8}\ttfamily,
    breakatwhitespace=true,
    breaklines=true,
    captionpos=b,
    keepspaces=true,
    numberstyle=\tiny\color{codegray},
    numbersep=2pt,
    showspaces=false,
    showstringspaces=false,
    showtabs=false,
    tabsize=1,
    linewidth=1.50\columnwidth,
    frame=tb,
    xrightmargin=0pt,
    xleftmargin=0.23cm,
    numbers=none,
    aboveskip=0.4cm,
    belowskip=0.4cm,
    keywordstyle=\color{blue},
    language=Scala,
    morekeywords={extension,using},
}

\lstdefinestyle{dataexample}{
    backgroundcolor=\color{backcolour},
    commentstyle=\color{codegreen}\rmfamily\itshape,
    keywordstyle=\color{keywords},
    stringstyle=\color{nightblue},
    basicstyle=\fontsize{7}{8}\ttfamily,
    breakatwhitespace=true,
    breaklines=true,
    captionpos=b,
    keepspaces=true,
    numberstyle=\tiny\color{codegray},
    numbersep=2pt,
    showspaces=false,
    showstringspaces=false,
    showtabs=false,
    tabsize=1,
    linewidth=1.65\columnwidth,
    frame=tb,
    xrightmargin=0pt,
    xleftmargin=0cm,
    numbers=none,
    aboveskip=0cm,
    belowskip=-0.2cm,
    keywordstyle=\color{blue},
    language=Scala,
    morekeywords={extension,using},
    morecomment=[l]{Step}
}

\newcommand{\nostepmethod}[0]{\textsc{Direct-Pred}}
\newcommand{\cotbaseline}[0]
{\textsc{CoT}}
\newcommand{\method}[0]{\textsc{DecInt}}

\newcommand{\ourdata}{\textsl{DeCU}}

\newif\ifshowchanges

\ifshowchanges
  \newcommand{\newremove}[1]{{\color{blue} \sout{#1}}}
  \newcommand{\newreplace}[2]{{\color{blue} \sout{#1}}{\color{red} #2}}
  \newcommand{\new}[1]{{\color{red} #1}}
\else
  \newcommand{\newremove}[1]{}
  \newcommand{\newreplace}[2]{{}{#2}}
  \newcommand{\new}[1]{{#1}}
\fi

\newcommand{\eg}{\emph{e.g.,}\ }
\newcommand{\ie}{\emph{i.e.,}\ }

\title{Natural Language Decomposition and Interpretation \\
 of Complex Utterances}

\author{%
Harsh Jhamtani \quad Hao Fang
\quad Patrick Xia \quad Eran Levy \\
\qquad {\bf Jacob Andreas}
\quad\ {\bf Ben Van Durme} \\
Microsoft Semantic Machines \ \texttt{<sminfo@microsoft.com>}
}

\begin{document}
\maketitle

\begin{abstract}

Designing natural language interfaces has historically required collecting supervised data to translate user requests into carefully designed intent representations. This requires enumerating and labeling a long tail of user requests, which is challenging. At the same time, large language models (LLMs) encode knowledge about goals and plans that can help conversational assistants interpret user requests requiring numerous steps to complete. We introduce an approach to handle complex-intent-bearing utterances from a user via a process of hierarchical natural language decomposition and interpretation. Our approach uses a pre-trained language model to decompose a complex utterance into a sequence of simpler natural language steps and interprets each step using the language-to-program model designed for the interface. To test our approach, we collect and release  \ourdata{}---a new NL-to-program benchmark to evaluate \textbf{De}composition of \textbf{C}omplex \textbf{U}tterances.\footnote{Our code and \ourdata{} dataset will be released.} Experiments show that the proposed approach enables the interpretation of complex utterances with almost no complex training data, while outperforming standard few-shot prompting approaches.
\end{abstract}

\section{Introduction}

Neural sequence models, pre-trained on large datasets of language and code, are extremely effective at parsing natural commands into programs, database queries, and other structured representations of user intent \cite{DBLP:journals/corr/abs-2107-03374,DBLP:conf/eacl/LiACGGM21,DBLP:conf/emnlp/ShinLTCRPPKED21, DBLP:journals/corr/abs-2206-10668}. 
However, developing an interface that enables a user to interact with a new API or software system still 
requires substantial system-specific data collection.
Users, meanwhile, may not be aware of the scope of this data collection, and pursue an open-ended set of goals -- including goals more complicated than those anticipated by system designers.

\begin{figure*}[t]
    \centering
   \includegraphics[width=.99\textwidth,clip,trim=0.4in 0.5in 0.5in 0.0in]{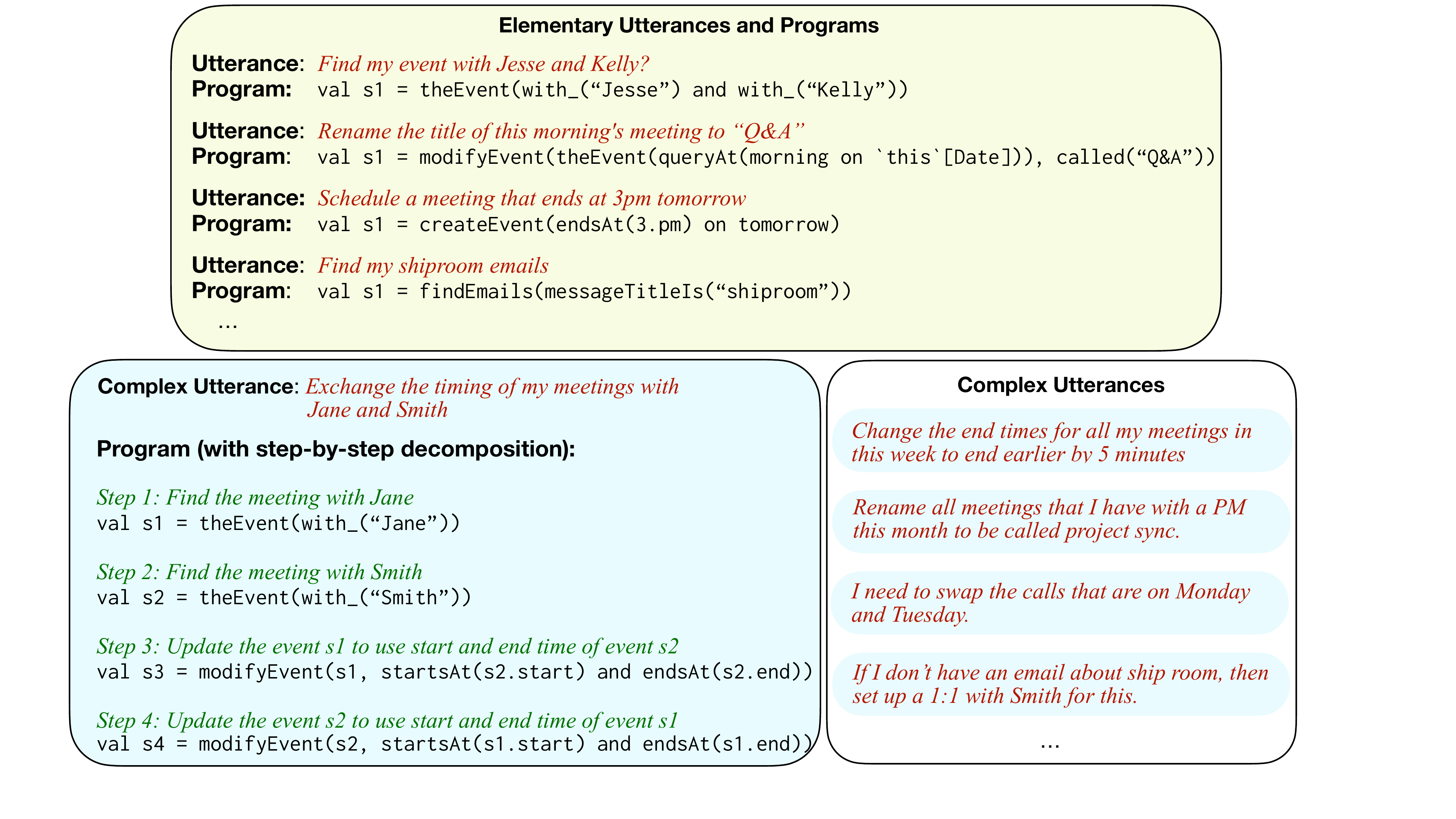}
    \caption{Parsing NL user utterances into programs.
    We study a scenario in which a large number of elementary utterances have been annotated with programs (top block),
    and we wish to build a model that can generalize to complex utterances (bottom blocks) requiring more elaborate programs.
    We introduce a method called \method{} that uses an LLM to decompose a {\color{complexred}\textit{complex utterance}} by predicting {\color{stepgreen}\textit{simpler NL steps}},
    each of which is parsed to a \texttt{program} according to the annotated elementary utterances.}
    \label{fig:pull}
     
\end{figure*}

\newreplace{
Recent work has demonstrated that large language models (LLMs) encode knowledge that can be used to interpret complex user goals requiring numerous steps to complete, in setups such as question answering \citep{DBLP:journals/tacl/WolfsonGGGGDB20, DBLP:journals/corr/abs-2210-02406} and  embodied agents  \cite{ahn-etal-2022-do,huang-etal-2022-language}.
Unlike such past work, we are concerned with generating programs in a carefully designed intent representation. %
In this paper, we present \method{}, a novel approach to decompose complex utterances into a sequence of simpler NL steps,
each resembling a simpler elementary utterance that an existing language-to-program parser for the NL interface can map to a sub-program. Consider the utterance \textit{``Exchange the timing of my meetings with Jane and Smith''} (\autoref{fig:pull}).
\method{} breaks the utterance down into four NL steps, using a pre-trained LLM and just a few annotated decompositions. 
The generated NL steps are parsed into programs, relying primarily on a larger set of existing elementary utterances associated with simpler programs in the target representation. 
\method{} thus enables an NL interface system to handle user requests representing complex goals (never seen by a semantic parser) by breaking them into a series of NL steps that are interpreted into APIs (never seen by an LLM).}
{
In this paper, we present \method{}, an approach to \textbf{dec}ompose complex utterances into a sequence of simpler NL steps,
each resembling a simpler elementary utterance that an existing language-to-program \textbf{int}erpreter for the NL interface can parse to a sub-program. Consider the utterance \textit{``Exchange the timing of my meetings with Jane and Smith''} (\autoref{fig:pull}).
\method{} breaks the utterance down into four NL steps, using a pre-trained LLM and just a few annotated decompositions. 
The generated NL steps are parsed into programs, relying primarily on a relevant (to the step being parsed) subset of 
a larger set of existing elementary utterances associated with simpler programs in the target representation. 
\method{} thus enables an NL interface system to handle user requests representing complex goals (never seen by a semantic parser) by breaking them into a series of NL steps that are interpreted into APIs (never seen by an LLM).
Our work is related to recent work which demonstrates that large language models (LLMs) encode knowledge that can be used to interpret complex user goals requiring numerous steps to complete, in setups such as question answering \citep{DBLP:journals/tacl/WolfsonGGGGDB20, DBLP:journals/corr/abs-2210-02406} and embodied agents  \cite{ahn-etal-2022-do,huang-etal-2022-language}. 
Compared to such past work, we are concerned with generating programs in a carefully designed intent representation. Starting with labeled elementary utterances, we wish to be able to parse complex utterances that are broader in scope compared to the abundant elementary utterances. %
}

To study utterance decomposition in %
NL-to-program space, 
we collect and release \ourdata{}---a new benchmark dataset to evaluate models for \textbf{De}composition of \textbf{C}omplex \textbf{U}tterance. \ourdata{} consists of 
(1) a set of elementary utterances and corresponding programs for managing calendar events and emails
and (2) a diverse set of complex user utterances annotated with decompositions into sequences of elementary utterances and their corresponding program fragments.
Experiments on \ourdata{} show that \method{} outperforms direct few-shot prompting approaches, making it possible to build NL interfaces that accomplish complex goals without large amounts of complex labeled data.

\section{Task Overview}
We study the problem of parsing an NL user utterance $x$ into a program $y$ that correctly reflects user intent (\autoref{fig:pull}). 
We focus on a version of the problem with the following characteristics:
\begin{itemize}
    \item A domain developer has already collected a dataset of \textbf{elementary utterances} annotated with corresponding programs. These utterances represent narrow user goals associated with simple and short programs. %
    \item At test time, the system must interpret \textbf{complex utterances}. Such utterances require longer programs representing much broader user goals.
    \item For a small number of complex utterances, we have access to annotations consisting of both natural language decompositions into elementary utterances, and program annotations for elementary utterances. 
\end{itemize}

Annotated complex utterances will in general cover only a small part of the space of possible user requests, and our goal is to build a language-to-program model that can generalize to requests of very different kinds (\autoref{fig:pull}).

\lstset{style=dataexample}
\begin{figure*}[t]
    \centering
    \footnotesize
    \begin{tabular}{>{\centering\arraybackslash}m{2cm} m{12cm}}
        \toprule
        \textbf{Utterance 1:} & \textit{Change my meetings with Abby and those with Dan this week to start 5 minutes later.} \\
\textbf{Decomposition:} & 
\begin{lstlisting}[language=Scala,aboveskip=0cm,belowskip=-0.2cm,xleftmargin=0cm,style=dataexample]
Step 1: Find events with Abby this week
val s1 = findEvents(with_("Abby") and queryAt(`this`[Interval[Date]] and isWeek))
Step 2: Find events with Dan and without Abby this week
val s2 = findEvents(with_("Dan") and not(with_("Abby")) and queryAt(`this`[Interval[Date]] and isWeek))
Step 3: Set all meetings from the list of events s1 to start 5 minutes later
val s3 = s1.map((x: Event) => modifyEvent(x, startsAt(x.start.local.time + 5.minutes)))
Step 4: Set all meetings from the list of events s2 to start 5 minutes later
val s4 = s2.map((x: Event) => modifyEvent(x, startsAt(x.start.local.time + 5.minutes))) \end{lstlisting} \\
\midrule
\textbf{Utterance 2:} & \textit{Decline any meeting invitations that are scheduled during my weekly team meeting.} \\
\textbf{Decomposition:} & 
\begin{lstlisting}[language=Scala,aboveskip=0cm,belowskip=-0.2cm,xleftmargin=0cm,style=dataexample]
Step 1: Find the event called "team meeting" that recurs weekly.
val s1 = theEvent(called("team meeting") and recurringWeekly)
Step 2: Find all events.
val s2 = findEvents0
Step 3: Filter events from list s2 to only include ones that intersect with event s1 that are not s1.
val s3 = s2.filter((x: Event) => x.interval.intersects(s1.interval) && x.id != s1.id)
Step 4: Decline events in the list s3.
val s4 = s3.map((x: Event) => respond(x, ResponseStatusType.declined)) \end{lstlisting} \\
        \bottomrule
    \end{tabular}
\caption{Examples of complex utterances in \ourdata{}. Each utterance is accompanied by decompositions consisting of a sequence of NL steps and associated program fragments, annotated by domain experts.
}
    \label{tab:data-examples}
    \vspace{-6pt}
\end{figure*}

\section{Data}

Many existing relevant decomposition datasets focus on open-ended QA \cite{DBLP:journals/tacl/WolfsonGGGGDB20,DBLP:conf/naacl/KhotKRCS21, DBLP:journals/corr/abs-2210-02406} or robotics domains with a relatively small number of fixed allowed actions \cite{puig2018virtualhome,shridhar2020alfred}. %
By contrast, we are interested in the task of parsing a user utterance to a program  that represents the actions to be taken by the interface, grounded on a large number of fixed APIs. Moreover, we want to study how complex user utterances can be supported by the NL interface, without collecting a large amount of additional labeled data, by using decomposition in NL space.
To study such multi-step complex intent decomposition, we introduce a new dataset we call \ourdata{} (\textbf{De}composition of \textbf{C}omplex \textbf{U}tterances). 

The utterances in \ourdata{} focus on calendar events and emails.
The dataset contains both elementary utterances and complex utterances.
Elementary utterances (\S\ref{ssec:elementary})
are paired with declarative Scala3 programs based on a domain library (\S\ref{ssec:domain_library}) that admits a fixed set of APIs and specified types. 
Complex utterances (\S\ref{ssec:complex}) are annotated with a corresponding sequence of elementary utterances, each paired with a program. Only a few of these  complex utterances are included in the training set; they are mainly used to form a test set.

\autoref{fig:pull} illustrates an example, \textit{``Exchange the timing of my meetings with Jane and Smith''}. How such an utterance should be decomposed is domain-dependent:
here, the calendar API does not provide a single endpoint that can swap pairs of meetings; instead, the system must search for the two meetings individually, then update each of their times. \autoref{fig:pull} shows a possible decomposition into four steps. The first generated NL step, \textit{``Find the meeting with Jane''}, is translated to a program fragment: \texttt{val s1 = theEvent(with\_(\textrm{``}Jane\textrm{''}))}. Individual steps typically represent easier-to-solve inputs for the NL-to-program parser that primarily relies on the annotated elementary utterances.

In addition to domain-specific knowledge of APIs, decomposition of complex utterances often relies on domain-general reasoning and common sense knowledge -- for example, 
to avoid double-counting meetings that match two search results (\autoref{tab:data-examples}, utterance 1),
or to recognize that meetings cannot conflict with themselves (utterance 2).

\subsection{Domain Library}
\label{ssec:domain_library}

The domain library defines the set of types and functions available for program annotations.
Types model objects such as \texttt{Person} and \texttt{Event},
whereas functions represent actions that can be taken by the agent,
including high-level APIs (\eg \texttt{createEvent}, \texttt{findEmails}),
low-level operations (\eg \texttt{min}, \texttt{+}),
predicate constructors (\eg \texttt{called}, \texttt{startsAt}),
etc.
The domain library for \ourdata{} is packaged as standard Scala source code,
consisting of 33 types and over 200 functions.\footnote{%
Some built-in types (\eg \texttt{String}, \texttt{Boolean}), functions (\eg \texttt{map}), 
and control flow statements (\eg \texttt{if}) are not explicitly defined and counted. Appendix~\ref{appendix:domain_library} provides more details.
}

\begin{figure*}[t]
    \centering
    \includegraphics[width=.99\textwidth,clip,trim=0.4in 1.5in 1.4in 0.1in]{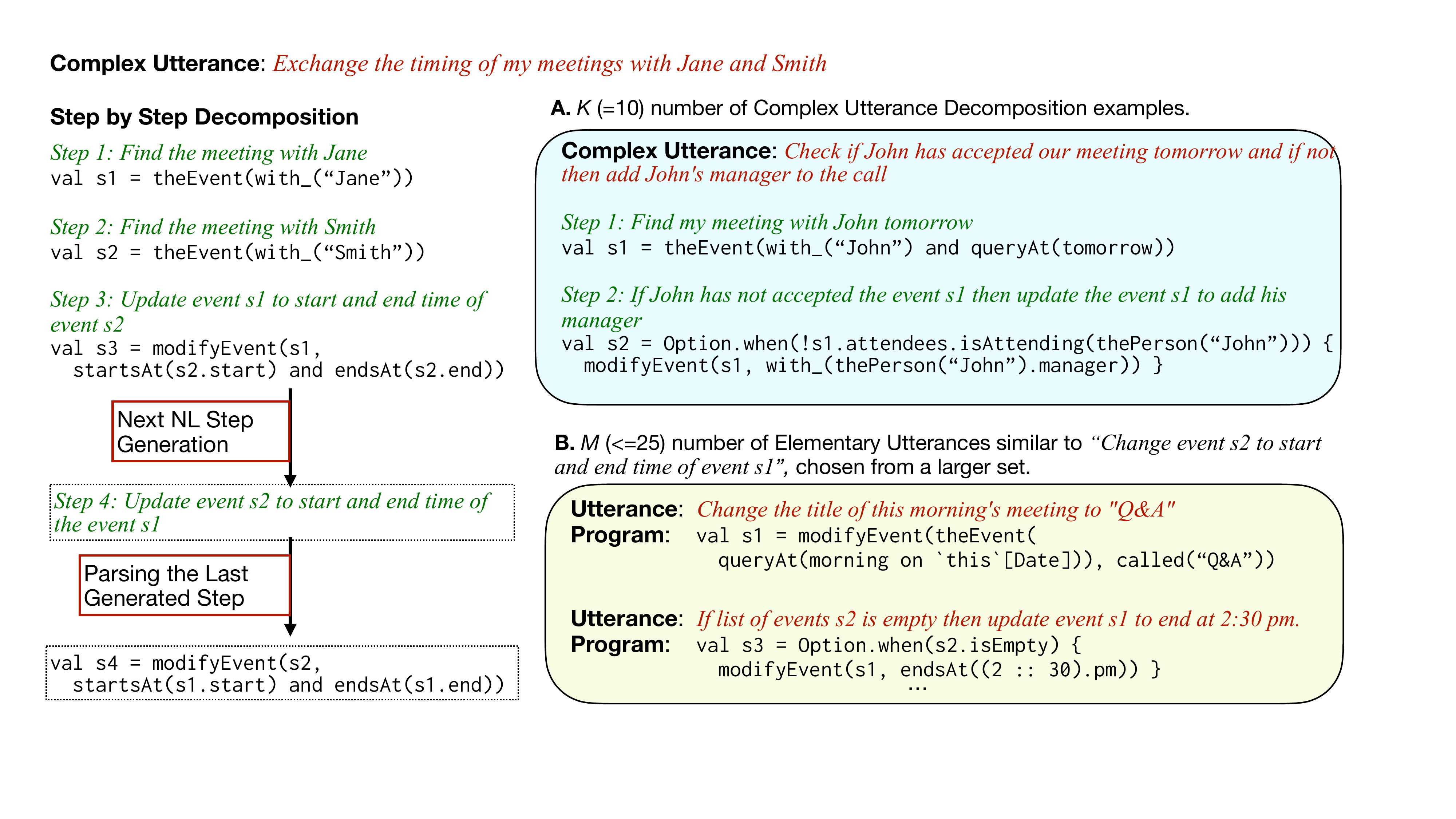}
    \caption{
    \method{} maps complex utterances into elementary steps, each of which is parsed in sequence to arrive at a final program.  %
    NL decomposition and program generation steps are interleaved. %
    While parsing a step, up to $M$ similar examples of elementary utterances are retrieved.
    }
    \label{fig:method}
    \vspace{-6pt}
\end{figure*}

\subsection{Elementary Utterances}
\label{ssec:elementary}
\ourdata{} contains 841 elementary utterances paired with programs.
A few examples are shown in the top box in \autoref{fig:pull}. %
These utterances are \emph{elementary} in that they represent narrow user goals such as creating or deleting a single meeting, which can typically be achieved using a single API.
As such, they have relatively short programs, generally less than 5 tokens.\footnote{\label{note1}To compute this statistic, programs are split into tokens based on heuristics, treating API names, argument names, and values as individual tokens.} 
Examples are written and reviewed by domain experts who are familiar with the domain library (on account of their experience from working with a deployed system leveraging such a library) and annotation guidelines.

\subsection{Complex Utterances}
\label{ssec:complex}
To study how \emph{complex} utterances can be supported by an NL interface,
we collect a diverse set of more involved user requests, and annotate these with decompositions into elementary steps, along with programs for each step. 
As the name suggests, compared to elementary utterances, these utterances represent more complex and broader user goals, with the corresponding programs typically being much longer (an average of 14.5 tokens per program).  
To collect complex utterances, we employ a mix of manual authoring and automated utterance generation. 
Manual authoring is performed by domain experts with a focus on diversity and goals that require the composition of multiple calls to the domain APIs.
For automated collection techniques, we generate utterances using GPT-3 \cite{brown2020language} prompted with a few random examples of manually-authored utterances. About 60\% of all the collected utterances were generated automatically.
Appendix \ref{appendix:data} provides more details on utterance collection.
Examples are shown in \autoref{fig:pull}.

\paragraph{Decomposition Annotations:}
Six annotators familiar with the domain (annotators had past experience working with the domain library) decompose complex utterances into elementary ones. 
When results from earlier steps must be reused, these NL decompositions may include explicit reference to earlier step outputs (\autoref{tab:data-examples}). %
More information about annotator instruction is provided in Appendix~\ref{appendix:data}.
Each annotation was additionally reviewed by two additional domain experts, separate from the set of 6 annotators.

\paragraph{Data Statistics:}
We collected a total of \newreplace{126}{210} unique complex utterances. \newreplace{each of which is paired with accompanying decomposition annotations.}{
The dataset is a mix of 126 utterances paired with annotated programs and 84 that are unannotated. As discussed later, in addition to reference-based metrics, we also provide various reference-less metrics that do not require annotations.}
While it is a relatively small count, note that most of the data (\newreplace{116 out of 126}{200 out of 210}) is used to construct an evaluation set, as we are interested in learning to generalize from very small numbers of training examples.
Annotated complex utterances in our full dataset exhibit a diverse range of properties (an utterance can have multiple): 55\% use a map operation (for-loop), 36\% contain actions based on a condition, 31\% use a filter operation, 24\% query about calendar/email, 37\% contain a create meeting action, 9\% contain a delete meeting action, and 31\% contain a modify meeting action.
The average number of decomposition steps in our data is 3, with a maximum of 7 steps. 
The average number of tokens in each program %
is 14.5\footnotemark,
while the average number of 
tokens in the program fragment corresponding to a single step is 4.8. 
For comparison, the average number of tokens in the programs for elementary utterances is 4.5.

\section{Approach}

The \method{} approach, illustrated in \autoref{fig:method}, maps a complex utterance $x$ to a sequence of interpretable lower-level NL steps ($z_1,z_2,..$) that resemble elementary utterances. Each step or low-level utterance $z_j$ is parsed into a program fragment $y_j$. In particular, \method{} maps from commands to programs according to the following iterative generative process:

\begin{enumerate}
\item
\textbf{Natural Language Decomposition}: \\ $z_j \sim p(\cdot \mid x,z_{<j},y_{<j})$.
\item 
\textbf{Program Generation}: \\ $y_j \sim p(\cdot \mid x,z_{\le j},y_{<j})$.
\end{enumerate}
NL Decomposition (\S\ref{sec:step_gen}) and program generation (\S\ref{sec:step_parse}) steps are interleaved, with later portions of the language decomposition conditioned on earlier program fragments. In principle, one could also condition on the return values of the earlier program fragments (see Limitations section).  We do not do so in this paper, as running the programs would require API implementations and input data.

\subsection{Natural Language Decomposition}
\label{sec:step_gen}
The NL decomposition stage generates the next NL step $z_j$ conditioned on the user utterance $x$ and any previously generated steps and program fragments. 
We obtain $z_j$ by greedy decoding from a pre-trained LLM in a few-shot in-context learning setup \cite{brown2020language}. 
The model is prompted with $K=10$ example decompositions, each of which consists of an
utterance $x$ followed by any previous steps and their program fragments, all concatenated together ($x,z_1,y_1,z_2,,...,z_N,y_N$). 
We additionally found it useful to include a list of up to $M$ elementary NL utterances at the start of the prompt (before the $K$ decomposition examples), selecting the ones with highest BM25 similarity to the input utterance. This is intended to inform the model about the kind of elementary steps the NL-to-program parser can handle.
(An example constructed prompt is shown in Appendix \ref{appendix:prompt_example}.) 
Example decompositions are taken from the set of 10 complex utterances in the training split of \ourdata{}.

\method{}'s ability to perform NL decomposition thus results from a combination of parametric knowledge about the structure of programs in general (the result of pretraining) and non-parametric knowledge about the domain of interest (obtained via in-context learning).
Together, these enable generalization to structurally novel user requests.
For example, there are no training examples that involve exchanging the timing of two meetings (the test example in \autoref{fig:method}), but \method{} nonetheless synthesizes a correct program.

\subsection{Program Generation}
\label{sec:step_parse}

The program generation step synthesizes a program fragment $y_j$ for a given NL step $z_j$, conditioned on any preceding steps and incomplete program. 
This is a well-studied semantic parsing problem, and we design the NL-to-program parser largely following past work that applies pre-trained LLMs.
We use in-context learning with dynamically selected prompt examples from the set of elementary examples data \cite{brown2020language}.  As before, we use greedy decoding.
Generated program fragments may refer to previously generated fragments using named step variables.
For a given NL utterance or step, we identify up to $M$ examples from the set of elementary utterances, where each example is an (utterance, program) pair (as shown in box B in \autoref{fig:method}). The selection of the examples is based on the similarity of the utterance to the intermediate NL step being parsed. To compute similarity, we again use BM25, as in past work \cite{DBLP:conf/naacl/RubinHB22, DBLP:journals/corr/abs-2206-10668}. 
In pilot experiments on training data, we discovered it was useful to also include the $K$ decomposition examples at the bottom of the prompt (detailed prompt example shown in Appendix \ref{appendix:prompt_example}). This may be because the decomposition examples provide a demonstration of how to generate program fragments for a step conditioned on previous steps and help bridge any possible domain shift from elementary to complex utterances.

\subsection{\new{Baselines}}
The \method{} method decomposes a complex utterance into NL steps, separately parsing each step, and using internal variable references to assemble a larger program.
The standard few-shot prompting approach for tasks like this one \citep[\eg][]{DBLP:journals/corr/abs-2206-10668}
instead directly predicts the parse without generating the intermediate NL steps.
We compare to this approach, which we denote \textbf{\nostepmethod{}},  in our experiments. There are a few key differences compared to the \method{} method.
Complex utterance examples are presented without the intermediate NL steps (\ie each utterance is paired with a multi-line program).
The output generation is a single-step process since there are no intermediate NL steps that need to be generated.
As with \method{}, examples of elementary utterances are also included in the prompt. 
\new{We also consider a \textbf{\textsc{CoT}} \cite{wei2022chain} baseline, wherein the model first predicts all intermediate NL steps and then predicts the program. Accordingly, the complex utterance examples in the prompt are annotated with intermediate steps. This baseline resembles the method proposed in \citet{DBLP:journals/corr/abs-2303-06689}. Note that compared to \textsc{CoT}, \method{} interleaves step generation and parsing, and dynamically updates the subset of exemplars from elementary utterances to be relevant to the step being parsed.}

\new{
We also report results using a variant of \method{}  that relies only on $K$ decomposition exemplars but without access to elementary utterances ($M$=0 instead of 25). We refer to such a baseline as \textbf{\textsc{Few-Shot}}. %
Finally, we also report results for a variant of \method{} that uses only a single decomposition exemplar ($K$=1 instead of 10), and thus relies almost entirely on the elementary utterances from the underlying domain. We refer to the variant 
as \textbf{\textsc{Elementary-Only}}.
}

\section{Experiments}

\subsection{Evaluation}

\paragraph{Overlap with Reference Programs:}
We report Exact Match (\textbf{EM}) and character-based edit distance (\textbf{CER}) metrics\footnote{https://huggingface.co/spaces/evaluate-metric/cer} against the gold program.  
Before computing these metrics, we normalize the programs by lowercasing the entire program and removing extra spaces.
Since there can be multiple possible ways to express the target multi-line program, Exact Match can only be viewed as a lower-bound metric for this task.
These metrics are reported only for the subset of the data that consists of annotated reference programs.

\paragraph{Well-formed Evaluation:}
Additionally, we report the fraction of predictions that are valid (\textbf{WellForm}) under the domain library,
\ie the full program follows correct syntax and only uses functions available in the library.
Note that WellForm does not necessarily represent correctness with respect to the user goal. We report the metric for the entire test set.

\paragraph{Program Correctness:}
Finally, we report the overall correctness of the generated programs. \new{We define a program to be correct overall if: it is well-formed, and correctly represents the user request.}
\newremove{as per human evaluators (\textbf{Correct}). 
The human evaluators are familiar with the domain library. They are presented with the user utterance, along with the program, and are asked to give a binary judgment rating on whether the program correctly represents the user intent. We report results on outputs from \method{}, \nostepmethod{} and \cotbaseline{}. Outputs that are not well-formed (WellForm) are automatically considered to be incorrect in this evaluation (but are included in the denominator for all evaluations).}
\new{
We use GPT-4 (gpt-4-32k) \cite{openai2023gpt4} to rate the correctness of the generated programs (\textbf{Correct}). The prompt consists of an instruction and four manually labeled exemplars (two ``correct'' and two ``incorrect'') followed by the test example. Each example is a user utterance followed by the associated program. The label is a natural language caption/explanation of the generated program, followed by a final verdict on whether either the generated program is ``correct'' or ``incorrect'' for the given user utterance -- following a chain-of-thought style prediction\footnote{The exact prompts used in Correct are presented in Appendix \ref{appendix:experiments}}.
\new{Since we have an automatic static analysis to infer exactly which programs are well-formed (WellForm)}, outputs that are not well-formed are automatically considered to be incorrect as per the definition above (but are included in the denominator for all evaluations). Note that the Correct metric is reference-less, is easier to scale than human evaluations, and correlates well with human ratings (Section \ref{ssec:programeval}). }

\subsection{Setup}

We consider the task of parsing complex utterances in \ourdata{} given only ten complex utterances (annotated with decompositions) to be used as training data (exemplars for in-context learning). We report results on the test set consisting of the remaining \newreplace{116}{200} complex utterances.
We use a maximum of $M\leq25$ additional elementary utterances (as many as permitted by the LM's context window) selected according to BM25 similarity with the step being parsed. 
We use OpenAI's \emph{text-davinci-003} model as the LLM for generating each NL step as well as for parsing it into a program.

\begin{table}[t]
    \centering
    \footnotesize
    \begin{tabular}{l@{\hskip 0.01in}c@{\hskip 0.03in}c@{\hskip 0.04in}c@{\hskip 0.04in}c}
        \toprule
    \textbf{System} & \textbf{Correct}$\uparrow$ & \textbf{WellForm}$\uparrow$ & \textbf{EM}$\uparrow$ & \textbf{CER}$\downarrow$ \\ 
    \midrule 
    \nostepmethod{} & $0.34$ & $0.36$ & $0.04$ & $0.44$ \\
    \new{\cotbaseline{}} & $0.25$ & $0.29$ & $0.05$ & $0.46$ \\
    \new{\textsc{Few-Shot}} & $0.13$ & $0.19$ & $0.00$ & $0.50$ \\ %
    \new{\textsc{Elementary-Only}} & $0.23$ & $0.31$ & $0.04$ & $0.54$ \\ %
    \method{} & $\bf0.41$ & $\bf0.46$ & $0.05$ & $\bf0.40$ \\
    \bottomrule
    \end{tabular}
    \caption{Quality of the generated program for complex utterances under various automated metrics. \newremove{$M$ represents the number of exemplars for the parser in the prompt, sourced from elementary utterances. $K$ represents the number of annotated complex utterance examples in the prompt.} %
    }
    \label{tab:auto_results}
\end{table}

\begin{figure*}[t]
    \centering
    \footnotesize
    \begin{tabular}{>{\centering\arraybackslash}m{2cm} m{12cm}}
    \toprule

        \textbf{Utterance 1:}  
    & \textit{Make me a meeting at 3 PM mountain time and refuse any meetings that happen at the same time.} \\ 
    
    \textbf{\method{}}
    & \begin{lstlisting}[language=Scala,aboveskip=0cm,belowskip=-0.2cm,xleftmargin=0cm,style=dataexample]
Step 1: Find events at 3 pm mountain time
val s1 = findEvents(queryAt(3.pm inZone TimeZone("MST"))) 
Step 2: Refuse any events in the list s1
val s2 = s1.map(x => respond(x, withResponse(ResponseStatusType.declined)))
Step 3: Create a meeting at 3 pm mountain time
val s3 = createEvent(createAt(3.pm inZone TimeZone("MST"))) 
\end{lstlisting} \\

    \textbf{\nostepmethod{}}
    & \begin{lstlisting}[language=Scala,aboveskip=0cm,belowskip=-0.2cm,xleftmargin=0cm,style=dataexample]
val s1 = createEvent(createAt(3.pm inZone TimeZone("MST"))) 
val s2 = findEvents(queryAt(3.pm inZone TimeZone("MST"))) 
val s3 = s2.map(x => refuseEvent(x))
\end{lstlisting}  \\

    \textbf{\cotbaseline{}}
    & \begin{lstlisting}[language=Scala,aboveskip=0cm,belowskip=-0.2cm,xleftmargin=0cm,style=dataexample]
Step 1: Create a meeting at 3 pm mountain time
Step 2: Refuse any meetings that happen at the same time"}
val s1 = createEvent(createAt(3.pm inZone TimeZone(\"MST\")))
val s2 = findEvents(queryAt(3.pm inZone TimeZone(\"MST\"))).map(x => refuseEvent(x))
\end{lstlisting}  \\
    
    \textbf{Comment} 
    & We observe that hallucination of functions is much more prominent in the baselines \nostepmethod{} and \cotbaseline{},
    as can be seen in the use of the non-existing function \texttt{refuseEvent} \\

    \midrule

    \textbf{Utterance 2:} & \textit{I'd like to divide the meeting at noon in two, so split off the last fifteen minutes as a separate event.} \\
    
    \textbf{\method{}} 
    & \begin{lstlisting}[language=Scala,aboveskip=0cm,belowskip=-0.2cm,xleftmargin=0cm,style=dataexample]
Step 1: Find the event at noon
val s1 = theEvent(createAt(noon)) 
Step 2: Subtract 15 minutes from the end time of the event s1
val s2 = s1.end - 15.minutes 
Step 3: Update the event s1 to end at the time instant s2
val s3 = modifyEvent(s1, endsAt(s2)) 
Step 4: Create a new event starting at the time instant s2
val s4 = createEvent(startsAt(s2))
\end{lstlisting} \\

    \textbf{Comment} 
    & Output program from \method{} was judged as incorrect since it doesn't specify the duration of the second event. \\

    \bottomrule

    \end{tabular}
    
    \caption{Outputs from \method{} and baselines on sample complex utterances in \ourdata{}.}
    \label{tab:qual-examples}
\end{figure*}

\subsection{\new{Evaluation of Generated Program}}
\label{ssec:programeval}

\autoref{tab:auto_results} reports various automated \newreplace{evaluations and human evaluations}{metrics}. 
\method{} outperforms \newreplace{\nostepmethod{}}{all the baselines}, sometimes by a wide margin. 
As can be seen in the table, \method{} outputs receive an overall correctness score (\textbf{Correct}) of 41\% for complex utterances compared to 34\% and 25\% for the baselines \nostepmethod{} and \cotbaseline{} respectively.\footnote{The differences are significant ($p < 0.05$) using bootstrap resampling.} We posit that \method{} is able to make more effective use of pretraining by breaking down a complex command into NL steps and retrieving relevant exemplars for each step. 
\new{Further, \textsc{Few-Shot}, that is equivalent to \method{} with M=0, fares badly, suggesting that \method{} relies on information from elementary utterances in addition to supervised decompositions. 
Finally, \textsc{Elementary-Only}, which is equivalent to \method{} with K=1, also does worse than \method{}, suggesting the usefulness of a handful of supervised decompositions.}
Note, however, that a \newreplace{53\% of the predictions from \nostepmethod{} and 43\% from \method{}}{54\% of the predictions from \method{}} are not well-formed, indicating that even structural generalization in \ourdata{} remains a major challenge. \new{Nonetheless, \method{} fares better compared to other methods on WellForm metric.}

\newremove{To further characterize \method{}, 
we carry out two ablation studies (results in \autoref{tab:auto_results}): 
(1) \method{}(M=0) without using elementary utterances ($M=0$ instead of 25) and
(2) \method{}(K=1) when using only 1 decomposition example ($K=1$ instead of 10).
As we expected, both the variants get worse scores than \method{} on all the automated metrics. Eliminating elementary utterances produces a large drop, suggesting that \method{} relies on information about the distribution of elementary utterances in addition to supervised decompositions.}

\new{
\noindent \textbf{Human Evaluation for Program Correctness:}
We also obtained the overall program correctness rating (``correct" vs ``incorrect" for a user utterance) from human evaluators familiar with the domain library. Just as was the case with \textbf{Correct} metric, outputs that are not well-formed are automatically considered incorrect. The aggregate scores for \method{}, \nostepmethod{} and \cotbaseline{} (our method and the two top performing baselines as per automated \textbf{Correct} metric) under human evaluation are 41\%, 33\% and 26\% respectively, which are very close to the scores for these methods under the automated \textbf{Correct} metric. 
Additionally, we observe a high correlation between human annotator-provided judgment and \textbf{Correct} judgments (a more detailed correlation analysis is provided in the Appendix \ref{appendix:experiments}).} %

\noindent \textbf{Results with \newreplace{another LLM (GPT-4)}{other LLMs}:}
\new{We also report results using GPT-4 (\emph{gpt-4-32k}) \new{and LLAMA-2-70B \cite{touvron2023llama}} as the underlying LLM. Due to cost considerations, we report results only for the top three methods from \autoref{tab:auto_results}. We observe that \method{} outperforms the baselines, demonstrating that the proposed approach is effective across underlying LLMs (\autoref{tab:gpt4results}).}

\begin{table}[th]
    \centering
    \footnotesize
    \begin{tabular}{lcc} %
        \toprule
        \textbf{System} & \multicolumn{2}{c}{\textbf{Correct}$\uparrow$/  \textbf{WellForm}$\uparrow$} \\ %
        & \textbf{GPT-4} & \textbf{LLAMA2-70B} \\
        \midrule 
         \nostepmethod{} & $0.35$ / $0.39$ & $0.25$ / $0.41$ \\ %
         \textsc{CoT} & $0.37$ / $0.40$ & $0.23$ / $0.32$ \\ %
         \method{} & $\bf0.49$ / $\bf0.56$ & $\bf0.35$ / $\bf0.50$ \\ %
        \bottomrule
    \end{tabular}
    \caption{Results using GPT-4\new{/LLAMA-2-70B} as the underlying LLM.
    }
    \label{tab:gpt4results}
    \vspace{-3pt}
\end{table}

\subsection{Evaluation of NL Decomposition}

\newremove{We conduct two levels of analysis on the NL steps  in \method{} outputs and the canonical decomposition steps written by domain experts for all 116 examples in the test set.

\noindent \textbf{Overall Evaluation:}}
We measure whether the NL decomposition steps altogether are \emph{sufficient and correct} to complete the user request.\footnote{Unless stated otherwise, all analysis uses outputs with \emph{text-davinci-003} as the underlying LLM} 
For example, the output from \method{} for the second utterance in \autoref{tab:qual-examples} is \emph{not} sufficient and correct because the fourth step fails to specify the duration of the meeting, which is supposed to be 15 minutes as per user request.
A random subset of 40 of \method{} NL predictions and corresponding expert annotations were manually labeled by one of the authors as correct or incorrect. The expert annotations and \method{} predictions were rated as 98\% and 85\% correct, respectively. 
Future work can explore ways to further improve the accuracy of the predicted NL steps. %
\new{We also conducted a step-level evaluation, which we discuss in \autoref{appendix:experiments}.}

\newremove{During system development, we also explored an automatic evaluation based on GPT-3 \cite{brown2020language}. The prompt consists of an instruction and 6 manually labeled exemplars followed by a test example. %
The NL steps predicted by \method{} are rated as correct and sufficient about $90\%$ of the time as per this automatic metric.
We then compared the manual labels and the automatic ones, and found %
Cohen's Kappa coefficient between them is 0.28, demonstrating a fair correlation (more details in \autoref{appendix:experiments}).}

\newremove{
\noindent \textbf{Step-by-Step Evaluation:}
Following ROSCOE \cite{golovneva-etal-2023-roscoe},\footnote{
We adopted ROSCOE as much as possible, though we received some feedback 
from crowdworkers that some dimensions (such as factuality, hallucination, and coherency) were difficult to judge for our evaluation task.}
we evaluate the quality of individual steps on 9 dimensions: grammar, factuality, hallucination, redundancy, repetition, missing information, coherency, commonsense, and arithmetic.
We recruited annotators from  Amazon Mechanical Turk to provide binary classification ratings (\textit{yes} or \textit{no}) for each step in the decomposition on all 9 dimensions.
For each question, 3 judgments are collected and the majority-voted answer is used as the final judgment.
For quality control, we restrict to annotators located in the United States or Canada
who have an approval rate higher than $85\%$
and have successfully solved a qualification task where we match their answers on the same set of questions against answers manually annotated by one of the authors. We pay 0.25 USD per example per question.
The step-by-step evaluation results shown in the bottom block of \autoref{tab:mturk_eval} 
suggest that the quality of individual steps is very high.
\method{}-predicted steps are rated similar to expert steps on almost all dimensions,
except on the ``missing information'' dimension, where the gap is noticeable.
Note that it is possible for all steps in a program to be judged individually correct, but fail to complete the user request. } %

\subsection{Qualitative Analysis} %
We provide example predictions in \autoref{tab:qual-examples}, with additional examples provided in \autoref{tab:qual-examples-appendix} and \autoref{tab:qual-examples-appendix-errors} in the Appendix.
Additionally, we perform an error analysis of the NL-to-program step of \method{}. %
We restrict the study to the predictions that were labeled as incorrect in \autoref{tab:auto_results}.  The most common issues are those that make the program not well-formed, as summarized in \autoref{tab:auto_results}. 
Many errors are due to \emph{nonexistent APIs / API arguments} (21\% of the incorrect programs have at least this problem) and \emph{nonexistent type attribute} (43\%).
A smaller number result from even more basic \emph{syntax errors} and \emph{type mismatches} (17\%).
Future work could constrain the outputs of the parser \cite{DBLP:conf/emnlp/ShinLTCRPPKED21} to only use allowed functions and follow correct syntax, 
though such approaches can substantially increase the cost of decoding.

A few errors result from predictions that capture only 
\emph{partial user intent} (6\%). For example, for utterance 2 in \autoref{tab:qual-examples}, the prediction does not capture the user intent of creating the second event for 15 minutes. 
Many of the remaining errors involve more fundamental semantic mismatches between user intents and model outputs.
For example, for \textit{``Loop around all my 1/1 meetings this week so that they also happen next week''}, the prediction updates the meetings this week instead of creating another set of meetings next week.

\newremove{Evaluation on Elementary Utterances: }
\newremove{To contextualize model performance on complex utterances, we conclude by analyzing
how the NL-to-program semantic parser fares on elementary utterances in the \ourdata dataset.
We split the elementary utterances data, consisting of 841 utterances, into train, dev and test splits in the ratio 70:15:15. 
The NL-to-program parser used in the experiment is the same as the one described earlier for the complex utterances -- we use in-context learning with large language models, with dynamically selected prompt examples from the train split.\footnote{%
Note that this experiment is limited to elementary utterances and \emph{no} complex utterance decompositions are used.}
We observe $0.60$ exact match (EM) and $0.10$ character edit rate (CER). 
While programs that fail exact match can potentially still be correct, these
results suggest that the underlying NL-to-program method itself has room for improvement.
We provide additional experiment details and analysis in Appendix \ref{appendix:experiments}.
}

\section{Related Work}

Past work has explored using 
command decomposition to break down complex tasks or requests into smaller subtasks that are easier to manage. 
The LaMDA model \cite{DBLP:journals/corr/abs-2201-08239}, for example, is capable of breaking down ``How to'' type queries into steps.
However, generated steps are not tied to any actions or APIs, and are more in the form of a narrative rather than executable steps.

\citet{DBLP:conf/naacl/KhotKRCS21} decompose a question into sub-questions that can be answered by a neural factoid single-span QA model and a symbolic calculator. %
\citet{DBLP:journals/corr/abs-2209-15003} decompose an utterance using a syntactic parse. However, not all utterances in our dataset would lend to such a style of decomposition, since all required actions might not align to a part of the parse. Recent work \cite{DBLP:journals/corr/abs-2303-06689} has also explored first generating an entire plan in NL and then generating a program. \newremove{Our setup interleaves step and program generation, and involves the interpretation of steps into a domain-specific representation as opposed to generating Python programs.} %
\new{\citet{paranjape2023art} focus on using tools and python scripts to complete a given task such `Translate into Pig Latin'. %
Compared to such past work, the complex utterances in our case are decomposed into intermediate steps that are parsed into a sub-program in the target representation as opposed to generating Python programs. Additionally, these sub-programs are a part of the final program output and thus we care about the accuracy of intermediate steps as well. 
}

A related area of research involves grounding high-level tasks, expressed in natural language, to a chosen set of actionable steps that a robot could take \cite{sharma22sl3, DBLP:journals/corr/abs-2209-11302,ahn-etal-2022-do,huang-etal-2022-language}. 
\citet{huang-etal-2022-language} propose a method to ground high-level tasks such as `make breakfast' to a set of actionable steps such as `open fridge'. %
Such work typically assumes a fixed inventory of low-level actions, and may not directly apply to setups like ours that additionally concerns with the interpretation of the steps into a rich target domain representation.

\section{Conclusion}
We have presented \method{}, an approach for interpreting complex user utterances by decomposing them into elementary natural language steps.
To evaluate methods for generating programs from natural language requests, we have introduced the \ourdata{} dataset, featuring a diverse set of utterances requiring substantial generalization from a small training set.
Experiments on \ourdata{} show that \method{} outperforms a standard few-shot prompting approach to program generation, with additional analysis revealing opportunities for improvement in both natural language decomposition and program generation phases.

\section*{Limitations}
The approach described in this paper does not condition on \emph{execution} results from intermediate steps, only generated programs themselves. 
Incorporating execution would improve the potential expressiveness of the model (\eg by allowing it to implement control flow operations conditioned on program results or exceptions). Program results might themselves be natural language strings (e.g., reminders or search results), enabling future extensions of \method{} to support an even richer space of requests.
We used pre-trained large-language models from OpenAI, through paid API access that may not be available for everyone or in the future. \new{However, our experiments using LLAMA-2-70B model \cite{touvron2023llama} should be easily reproducible.}
\new{We report and discuss several evaluation measures to check the quality of the predictions. One could also examine the outcome and side effects from executing the programs. However, lots of the queries require setting up a populated database and the outcome would vary as we execute the programs in different sandbox environments. Developing an evaluation setup with sandbox executions is challenging and remains an open research question.}

\paragraph{Ethics Statement:}
We leverage pre-trained neural language models such as GPT-3, and systems built using our approach might inherit some biases present in these pre-trained models. 
We build a system for NL-to-program, that users can leverage to command various NL interfaces. Such systems are not perfectly accurate and should be carefully deployed since they may lead to unintended side effects.

\bibliography{acl}

\begin{thebibliography}{25}
\expandafter\ifx\csname natexlab\endcsname\relax\def\natexlab#1{#1}\fi

\bibitem[{Ahn et~al.(2022)Ahn, Brohan, Brown, Chebotar, Cortes, David, Finn, Gopalakrishnan, Hausman, Herzog, Ho, Hsu, Ibarz, Ichter, Irpan, Jang, Ruano, Jeffrey, Jesmonth, Joshi, Julian, Kalashnikov, Kuang, Lee, Levine, Lu, Luu, Parada, Pastor, Quiambao, Rao, Rettinghouse, Reyes, Sermanet, Sievers, Tan, Toshev, Vanhoucke, Xia, Xiao, Xu, Xu, and Yan}]{ahn-etal-2022-do}
Michael Ahn, Anthony Brohan, Noah Brown, Yevgen Chebotar, Omar Cortes, Byron David, Chelsea Finn, Keerthana Gopalakrishnan, Karol Hausman, Alexander Herzog, Daniel Ho, Jasmine Hsu, Julian Ibarz, Brian Ichter, Alex Irpan, Eric Jang, Rosario~Jauregui Ruano, Kyle Jeffrey, Sally Jesmonth, Nikhil~J. Joshi, Ryan Julian, Dmitry Kalashnikov, Yuheng Kuang, Kuang{-}Huei Lee, Sergey Levine, Yao Lu, Linda Luu, Carolina Parada, Peter Pastor, Jornell Quiambao, Kanishka Rao, Jarek Rettinghouse, Diego Reyes, Pierre Sermanet, Nicolas Sievers, Clayton Tan, Alexander Toshev, Vincent Vanhoucke, Fei Xia, Ted Xiao, Peng Xu, Sichun Xu, and Mengyuan Yan. 2022.
\newblock \href {https://arxiv.org/abs/2204.01691} {Do as {I} can, not as {I} say: Grounding language in robotic affordances}.
\newblock \emph{arXiv: 2204.01691}.

\bibitem[{Brown et~al.(2020)Brown, Mann, Ryder, Subbiah, Kaplan, Dhariwal, Neelakantan, Shyam, Sastry, Askell et~al.}]{brown2020language}
Tom Brown, Benjamin Mann, Nick Ryder, Melanie Subbiah, Jared~D Kaplan, Prafulla Dhariwal, Arvind Neelakantan, Pranav Shyam, Girish Sastry, Amanda Askell, et~al. 2020.
\newblock \href {https://proceedings.neurips.cc/paper/2020/file/1457c0d6bfcb4967418bfb8ac142f64a-Paper.pdf} {Language models are few-shot learners}.
\newblock \emph{Advances in neural information processing systems}, 33:1877--1901.

\bibitem[{Chen et~al.(2021)Chen, Tworek, Jun, Yuan, de~Oliveira~Pinto, Kaplan, Edwards, Burda, Joseph, Brockman, Ray, Puri, Krueger, Petrov, Khlaaf, Sastry, Mishkin, Chan, Gray, Ryder, Pavlov, Power, Kaiser, Bavarian, Winter, Tillet, Such, Cummings, Plappert, Chantzis, Barnes, Herbert-Voss, Guss, Nichol, Paino, Tezak, Tang, Babuschkin, Balaji, Jain, Saunders, Hesse, Carr, Leike, Achiam, Misra, Morikawa, Radford, Knight, Brundage, Murati, Mayer, Welinder, McGrew, Amodei, McCandlish, Sutskever, and Zaremba}]{DBLP:journals/corr/abs-2107-03374}
Mark Chen, Jerry Tworek, Heewoo Jun, Qiming Yuan, Henrique~Ponde de~Oliveira~Pinto, Jared Kaplan, Harri Edwards, Yuri Burda, Nicholas Joseph, Greg Brockman, Alex Ray, Raul Puri, Gretchen Krueger, Michael Petrov, Heidy Khlaaf, Girish Sastry, Pamela Mishkin, Brooke Chan, Scott Gray, Nick Ryder, Mikhail Pavlov, Alethea Power, Lukasz Kaiser, Mohammad Bavarian, Clemens Winter, Philippe Tillet, Felipe~Petroski Such, Dave Cummings, Matthias Plappert, Fotios Chantzis, Elizabeth Barnes, Ariel Herbert-Voss, William~Hebgen Guss, Alex Nichol, Alex Paino, Nikolas Tezak, Jie Tang, Igor Babuschkin, Suchir Balaji, Shantanu Jain, William Saunders, Christopher Hesse, Andrew~N. Carr, Jan Leike, Josh Achiam, Vedant Misra, Evan Morikawa, Alec Radford, Matthew Knight, Miles Brundage, Mira Murati, Katie Mayer, Peter Welinder, Bob McGrew, Dario Amodei, Sam McCandlish, Ilya Sutskever, and Wojciech Zaremba. 2021.
\newblock \href {https://arxiv.org/abs/2107.03374} {Evaluating large language models trained on code}.
\newblock \emph{arXiv: 2107.03374}.

\bibitem[{Drozdov et~al.(2022)Drozdov, Sch{\"{a}}rli, Aky{\"{u}}rek, Scales, Song, Chen, Bousquet, and Zhou}]{DBLP:journals/corr/abs-2209-15003}
Andrew Drozdov, Nathanael Sch{\"{a}}rli, Ekin Aky{\"{u}}rek, Nathan Scales, Xinying Song, Xinyun Chen, Olivier Bousquet, and Denny Zhou. 2022.
\newblock \href {https://arxiv.org/abs/2209.15003} {Compositional semantic parsing with large language models}.
\newblock \emph{arXiv: 2209.15003}.

\bibitem[{Golovneva et~al.(2023)Golovneva, Chen, Poff, Corredor, Zettlemoyer, Fazel-Zarandi, and Celikyilmaz}]{golovneva-etal-2023-roscoe}
Olga Golovneva, Moya~Peng Chen, Spencer Poff, Martin Corredor, Luke Zettlemoyer, Maryam Fazel-Zarandi, and Asli Celikyilmaz. 2023.
\newblock \href {https://openreview.net/forum?id=xYlJRpzZtsY} {{ROSCOE}: A suite of metrics for scoring step-by-step reasoning}.
\newblock In \emph{Proceedings of the Eleventh International Conference on Learning Representations}.

\bibitem[{Huang et~al.(2022)Huang, Abbeel, Pathak, and Mordatch}]{huang-etal-2022-language}
Wenlong Huang, Pieter Abbeel, Deepak Pathak, and Igor Mordatch. 2022.
\newblock \href {https://proceedings.mlr.press/v162/huang22a.html} {Language models as zero-shot planners: Extracting actionable knowledge for embodied agents}.
\newblock In \emph{Proceedings of the 39th International Conference on Machine Learning}, volume 162 of \emph{Proceedings of Machine Learning Research}, pages 9118--9147. PMLR.

\bibitem[{Jiang et~al.(2023)Jiang, Dong, Wang, Shang, and Li}]{DBLP:journals/corr/abs-2303-06689}
Xue Jiang, Yihong Dong, Lecheng Wang, Qiwei Shang, and Ge~Li. 2023.
\newblock \href {https://arxiv.org/abs/2303.06689} {Self-planning code generation with large language model}.
\newblock \emph{arXiv: 2303.06689}.

\bibitem[{Khot et~al.(2021)Khot, Khashabi, Richardson, Clark, and Sabharwal}]{DBLP:conf/naacl/KhotKRCS21}
Tushar Khot, Daniel Khashabi, Kyle Richardson, Peter Clark, and Ashish Sabharwal. 2021.
\newblock \href {https://doi.org/10.18653/v1/2021.naacl-main.99} {Text modular networks: Learning to decompose tasks in the language of existing models}.
\newblock In \emph{Proceedings of the 2021 Conference of the North American Chapter of the Association for Computational Linguistics: Human Language Technologies, {NAACL-HLT} 2021, Online, June 6-11, 2021}, pages 1264--1279. Association for Computational Linguistics.

\bibitem[{Khot et~al.(2022)Khot, Trivedi, Finlayson, Fu, Richardson, Clark, and Sabharwal}]{DBLP:journals/corr/abs-2210-02406}
Tushar Khot, Harsh Trivedi, Matthew Finlayson, Yao Fu, Kyle Richardson, Peter Clark, and Ashish Sabharwal. 2022.
\newblock \href {https://arxiv.org/abs/2210.02406} {Decomposed prompting: {A} modular approach for solving complex tasks}.
\newblock \emph{arXiv: 2210.02406}.

\bibitem[{Li et~al.(2021)Li, Arora, Chen, Gupta, Gupta, and Mehdad}]{DBLP:conf/eacl/LiACGGM21}
Haoran Li, Abhinav Arora, Shuohui Chen, Anchit Gupta, Sonal Gupta, and Yashar Mehdad. 2021.
\newblock \href {https://doi.org/10.18653/v1/2021.eacl-main.257} {{MTOP:} {A} comprehensive multilingual task-oriented semantic parsing benchmark}.
\newblock In \emph{Proceedings of the 16th Conference of the European Chapter of the Association for Computational Linguistics: Main Volume, {EACL} 2021, Online, April 19 - 23, 2021}, pages 2950--2962. Association for Computational Linguistics.

\bibitem[{Novikova et~al.(2017)Novikova, Dusek, and Rieser}]{DBLP:conf/sigdial/NovikovaDR17}
Jekaterina Novikova, Ondrej Dusek, and Verena Rieser. 2017.
\newblock \href {https://doi.org/10.18653/v1/w17-5525} {The {E2E} dataset: New challenges for end-to-end generation}.
\newblock In \emph{Proceedings of the 18th Annual SIGdial Meeting on Discourse and Dialogue, Saarbr{\"{u}}cken, Germany, August 15-17, 2017}, pages 201--206. Association for Computational Linguistics.

\bibitem[{OpenAI(2023)}]{openai2023gpt4}
OpenAI. 2023.
\newblock \href {http://arxiv.org/abs/2303.08774} {Gpt-4 technical report}.

\bibitem[{Paranjape et~al.(2023)Paranjape, Lundberg, Singh, Hajishirzi, Zettlemoyer, and Ribeiro}]{paranjape2023art}
Bhargavi Paranjape, Scott Lundberg, Sameer Singh, Hannaneh Hajishirzi, Luke Zettlemoyer, and Marco~Tulio Ribeiro. 2023.
\newblock Art: Automatic multi-step reasoning and tool-use for large language models.
\newblock \emph{arXiv preprint arXiv:2303.09014}.

\bibitem[{Puig et~al.(2018)Puig, Ra, Boben, Li, Wang, Fidler, and Torralba}]{puig2018virtualhome}
Xavier Puig, Kevin Ra, Marko Boben, Jiaman Li, Tingwu Wang, Sanja Fidler, and Antonio Torralba. 2018.
\newblock Virtualhome: Simulating household activities via programs.
\newblock In \emph{Proceedings of the IEEE Conference on Computer Vision and Pattern Recognition}, pages 8494--8502.

\bibitem[{Roy et~al.(2022)Roy, Thomson, Chen, Shin, Pauls, Eisner, and Durme}]{DBLP:journals/corr/abs-2206-10668}
Subhro Roy, Sam Thomson, Tongfei Chen, Richard Shin, Adam Pauls, Jason Eisner, and Benjamin~Van Durme. 2022.
\newblock \href {https://arxiv.org/abs/2206.10668} {{BenchCLAMP}: {A} benchmark for evaluating language models on semantic parsing}.
\newblock \emph{arXiv: 2206.10668}.

\bibitem[{Rubin et~al.(2022)Rubin, Herzig, and Berant}]{DBLP:conf/naacl/RubinHB22}
Ohad Rubin, Jonathan Herzig, and Jonathan Berant. 2022.
\newblock \href {https://doi.org/10.18653/v1/2022.naacl-main.191} {Learning to retrieve prompts for in-context learning}.
\newblock In \emph{Proceedings of the 2022 Conference of the North American Chapter of the Association for Computational Linguistics: Human Language Technologies, {NAACL} 2022, Seattle, WA, United States, July 10-15, 2022}, pages 2655--2671. Association for Computational Linguistics.

\bibitem[{Sharma et~al.(2022)Sharma, Torralba, and Andreas}]{sharma22sl3}
Pratyusha Sharma, Antonio Torralba, and Jacob Andreas. 2022.
\newblock Skill induction and planning with latent language.
\newblock In \emph{Proceedings of the Annual Association for Computational Linguistics}.

\bibitem[{Shin et~al.(2021)Shin, Lin, Thomson, Chen, Roy, Platanios, Pauls, Klein, Eisner, and Durme}]{DBLP:conf/emnlp/ShinLTCRPPKED21}
Richard Shin, Christopher~H. Lin, Sam Thomson, Charles Chen, Subhro Roy, Emmanouil~Antonios Platanios, Adam Pauls, Dan Klein, Jason Eisner, and Benjamin~Van Durme. 2021.
\newblock \href {https://doi.org/10.18653/v1/2021.emnlp-main.608} {Constrained language models yield few-shot semantic parsers}.
\newblock In \emph{Proceedings of the 2021 Conference on Empirical Methods in Natural Language Processing, {EMNLP} 2021, Virtual Event / Punta Cana, Dominican Republic, 7-11 November, 2021}, pages 7699--7715. Association for Computational Linguistics.

\bibitem[{Shridhar et~al.(2020)Shridhar, Thomason, Gordon, Bisk, Han, Mottaghi, Zettlemoyer, and Fox}]{shridhar2020alfred}
Mohit Shridhar, Jesse Thomason, Daniel Gordon, Yonatan Bisk, Winson Han, Roozbeh Mottaghi, Luke Zettlemoyer, and Dieter Fox. 2020.
\newblock \href {https://openaccess.thecvf.com/content_CVPR_2020/papers/Shridhar_ALFRED_A_Benchmark_for_Interpreting_Grounded_Instructions_for_Everyday_Tasks_CVPR_2020_paper.pdf} {Alfred: A benchmark for interpreting grounded instructions for everyday tasks}.
\newblock In \emph{Proceedings of the IEEE/CVF conference on computer vision and pattern recognition}, pages 10740--10749.

\bibitem[{Singh et~al.(2022)Singh, Blukis, Mousavian, Goyal, Xu, Tremblay, Fox, Thomason, and Garg}]{DBLP:journals/corr/abs-2209-11302}
Ishika Singh, Valts Blukis, Arsalan Mousavian, Ankit Goyal, Danfei Xu, Jonathan Tremblay, Dieter Fox, Jesse Thomason, and Animesh Garg. 2022.
\newblock \href {https://arxiv.org/abs/2209.11302} {Progprompt: Generating situated robot task plans using large language models}.
\newblock \emph{arXiv: 2209.11302}.

\bibitem[{Thoppilan et~al.(2022)Thoppilan, Freitas, Hall, Shazeer, Kulshreshtha, Cheng, Jin, Bos, Baker, Du, Li, Lee, Zheng, Ghafouri, Menegali, Huang, Krikun, Lepikhin, Qin, Chen, Xu, Chen, Roberts, Bosma, Zhou, Chang, Krivokon, Rusch, Pickett, Meier{-}Hellstern, Morris, Doshi, Santos, Duke, Soraker, Zevenbergen, Prabhakaran, Diaz, Hutchinson, Olson, Molina, Hoffman{-}John, Lee, Aroyo, Rajakumar, Butryna, Lamm, Kuzmina, Fenton, Cohen, Bernstein, Kurzweil, y~Arcas, Cui, Croak, Chi, and Le}]{DBLP:journals/corr/abs-2201-08239}
Romal Thoppilan, Daniel~De Freitas, Jamie Hall, Noam Shazeer, Apoorv Kulshreshtha, Heng{-}Tze Cheng, Alicia Jin, Taylor Bos, Leslie Baker, Yu~Du, YaGuang Li, Hongrae Lee, Huaixiu~Steven Zheng, Amin Ghafouri, Marcelo Menegali, Yanping Huang, Maxim Krikun, Dmitry Lepikhin, James Qin, Dehao Chen, Yuanzhong Xu, Zhifeng Chen, Adam Roberts, Maarten Bosma, Yanqi Zhou, Chung{-}Ching Chang, Igor Krivokon, Will Rusch, Marc Pickett, Kathleen~S. Meier{-}Hellstern, Meredith~Ringel Morris, Tulsee Doshi, Renelito~Delos Santos, Toju Duke, Johnny Soraker, Ben Zevenbergen, Vinodkumar Prabhakaran, Mark Diaz, Ben Hutchinson, Kristen Olson, Alejandra Molina, Erin Hoffman{-}John, Josh Lee, Lora Aroyo, Ravi Rajakumar, Alena Butryna, Matthew Lamm, Viktoriya Kuzmina, Joe Fenton, Aaron Cohen, Rachel Bernstein, Ray Kurzweil, Blaise~Ag{\"{u}}era y~Arcas, Claire Cui, Marian Croak, Ed~H. Chi, and Quoc Le. 2022.
\newblock \href {http://arxiv.org/abs/2201.08239} {Lamda: Language models for dialog applications}.
\newblock \emph{CoRR}, abs/2201.08239.

\bibitem[{Touvron et~al.(2023)Touvron, Martin, Stone, Albert, Almahairi, Babaei, Bashlykov, Batra, Bhargava, Bhosale et~al.}]{touvron2023llama}
Hugo Touvron, Louis Martin, Kevin Stone, Peter Albert, Amjad Almahairi, Yasmine Babaei, Nikolay Bashlykov, Soumya Batra, Prajjwal Bhargava, Shruti Bhosale, et~al. 2023.
\newblock \href {https://arxiv.org/pdf/2307.09288.pdf} {Llama 2: Open foundation and fine-tuned chat models}.
\newblock \emph{arXiv preprint arXiv:2307.09288}.

\bibitem[{Wang et~al.(2016)Wang, Peter, Rosendahl, and Ney}]{wang-etal-2016-character}
Weiyue Wang, Jan-Thorsten Peter, Hendrik Rosendahl, and Hermann Ney. 2016.
\newblock \href {https://doi.org/10.18653/v1/W16-2342} {{C}harac{T}er: Translation edit rate on character level}.
\newblock In \emph{Proceedings of the First Conference on Machine Translation: Volume 2, Shared Task Papers}, pages 505--510, Berlin, Germany. Association for Computational Linguistics.

\bibitem[{Wei et~al.(2022)Wei, Wang, Schuurmans, Bosma, Xia, Chi, Le, Zhou et~al.}]{wei2022chain}
Jason Wei, Xuezhi Wang, Dale Schuurmans, Maarten Bosma, Fei Xia, Ed~Chi, Quoc~V Le, Denny Zhou, et~al. 2022.
\newblock Chain-of-thought prompting elicits reasoning in large language models.
\newblock \emph{Advances in Neural Information Processing Systems}, 35:24824--24837.

\bibitem[{Wolfson et~al.(2020)Wolfson, Geva, Gupta, Goldberg, Gardner, Deutch, and Berant}]{DBLP:journals/tacl/WolfsonGGGGDB20}
Tomer Wolfson, Mor Geva, Ankit Gupta, Yoav Goldberg, Matt Gardner, Daniel Deutch, and Jonathan Berant. 2020.
\newblock \href {https://doi.org/10.1162/tacl\_a\_00309} {Break it down: {A} question understanding benchmark}.
\newblock \emph{Trans. Assoc. Comput. Linguistics}, 8:183--198.

\end{thebibliography}

\clearpage
\appendix

\setcounter{table}{0}
\renewcommand{\thetable}{A\arabic{table}}

\setcounter{figure}{0}
\renewcommand{\thefigure}{A\arabic{figure}}

\section{Data Collection}
\label{appendix:data}

\subsection{Complex Utterance Collection}
The complex utterances in our data are collected by a mix of manual authoring and automated means described below: \\ 

\noindent \textbf{Utterances Authored by Expert Annotators:} A set of domain experts familiar with the elementary utterances are requested to author new complex utterances. 
They are informed of the following desiderata: 1) utterances should represent a more complex and broader intent compared to elementary utterances; 2) the set of utterances should be diverse.
To encourage creativity and diversity, the annotators were prompted with a set of keywords, and they are asked to author an utterance that spans at least some of the provided keywords. Similar approaches have been found useful in past work \cite{DBLP:conf/sigdial/NovikovaDR17}. The keywords are randomly sampled from a list of curated keywords, relevant to the calendaring and email domain. 
For each instance, we draw 5 keywords randomly from a much longer list of keywords constructed by the authors. Some of the keywords in our list are as follows: decline, pen, vacation, plan-my-day, project-sync, timezone, count-of-meetings, calendar-update, etc. 
\\

\noindent \textbf{Automatically Generated Utterances :} To scale the process of utterance collection and gather even more diverse utterances, we additionally generate complex utterances using GPT-3 \cite{brown2020language}, a pre-trained large language model. A few random examples of human-authored utterances are provided as in-context examples in the prompt, and new utterances are sampled. 
Specifically, we repeatedly sampled 10 utterances from the set of manually authored utterances to be used as prompt examples. We had additionally included an instruction `Now generate more utterances that are different from the above ones'. We sampled a new utterance with a temperature of 0.8.
About \newreplace{35\%}{60\%} of all the collected utterances were generated automatically via the described process.  \\

\noindent \textbf{Additional Information:}
Note that the utterances are limited to the English language, and expanding to other languages is a potential future extension. 
Additionally, note that the expert annotators were provided instructions that no personally identifiable information or offensive content should be present in the utterances. One of the authors also did a manual check of the collected data to ensure that the instructions were followed. 
All annotators were resident domain experts and were paid above the prevailing minimum page. 
The data annotators were provided with relevant information about the task and how the data would be used. Furthermore, the authors held an interactive session with the data annotators to give a brief overview and answer any questions.

\begin{figure*}
    \centering
    \includegraphics[width=.99\textwidth]{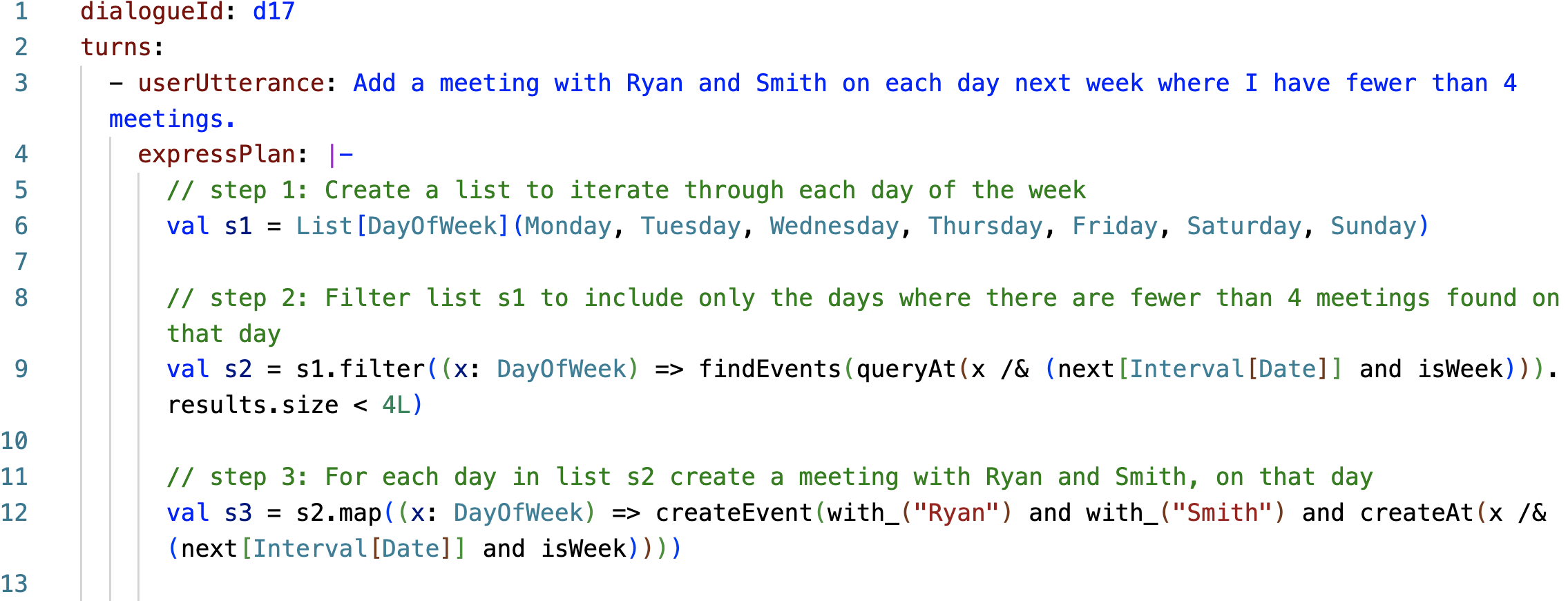}
    \caption{The interface used by domain experts to annotate the decompositions. The interface points out the syntax, type, and missing function errors, enabling the experts to author well-formed annotations. These ``express'' annotations are subsequently normalized by stripping comments, removing type annotations, code formatting, and selecting canonical function names when there is ambiguity (\texttt{concat} vs. \texttt{+}). The resulting programs conform to Scala syntax.}
    \label{fig:decomp_interface}
\end{figure*}

\subsection{Guidelines for Decomposition Annotations}
\label{appendix:guideline}
Expert annotators familiar with the domain are requested to annotate the decompositions of the utterances. 
\autoref{fig:decomp_interface} shows the interface used by the annotators. The interface automatically checks if the annotation is well-formed or not.
Annotators are asked to skip any utterances that cannot be annotated as per the domain library. 
We had a total of 6 domain experts who annotated the data. 
To ensure high quality, each annotation was additionally reviewed by at least 2 domain experts, separate from the set of 6 annotators.

Annotators are given guidance that to the extent possible, each step in decomposition is supposed to resemble an elementary utterance. Additionally, to refer to results from earlier steps, the results from $i^{th}$ can be referred to using variable $s_i$. An example can be seen in \autoref{fig:decomp_interface}.
We additionally provide guidance that NL steps should be grammatically correct full sentences. Moreover, annotators are advised to write the imperative commands using the present tense verb 
(\eg prefer \textit{``Filter reports \ldots''} over \textit{``Filtering reports \ldots''}).

\subsection{Decomposition Examples}

\lstset{style=dataexample}
\begin{figure*}[t]
    \centering
    \footnotesize
    \begin{tabular}{>{\centering\arraybackslash}m{2cm} m{12cm}}
        \toprule
        \textbf{Utterance 1:} & \textit{Change my meetings with Abby and those with Dan this week to start 5 minutes later.} \\
\textbf{Decomposition:} & 
\begin{lstlisting}[language=Scala,aboveskip=0cm,belowskip=-0.2cm,xleftmargin=0cm,style=dataexample]
Step 1: Find events with Abby this week
val s1 = findEvents(with_("Abby") and queryAt(`this`[Interval[Date]] and isWeek))
Step 2: Find events with Dan and without Abby this week
val s2 = findEvents(with_("Dan") and not(with_("Abby")) and queryAt(`this`[Interval[Date]] and isWeek))
Step 3: Set all meetings from the list of events s1 to start 5 minutes later
val s3 = s1.map((x: Event) => modifyEvent(x, startsAt(x.start.local.time + 5.minutes)))
Step 4: Set all meetings from the list of events s2 to start 5 minutes later
val s4 = s2.map((x: Event) => modifyEvent(x, startsAt(x.start.local.time + 5.minutes))) \end{lstlisting} \\
\midrule
\textbf{Utterance 2:} & \textit{For my reports who don’t already have a 1/1 call with me this week, schedule a 1/1 with each one of them.} \\
\textbf{Decomposition:} & 
\begin{lstlisting}[language=Scala,aboveskip=0cm,belowskip=-0.2cm,xleftmargin=0cm,style=dataexample]
Step 1: Retrieve my direct reports
val s1 = me.directReports
Step 2: Filter out the reports that do not have a one-on-one meeting with me this week
val s2 = s1.filter((x: Person) => findEvents(with_(x) and isOneOnOne and queryAt(`this`[Interval[Date]] and isWeek)).results.isEmpty)
Step 3: Create a one-on-one meeting with those filtered reports in list s2
val s3 = s2.map((x: Person) => createEvent(with_(x) and isOneOnOne)) \end{lstlisting} \\
\midrule
\textbf{Utterance 3:} & \textit{Decline any meeting invitations that are scheduled during my weekly team meeting.} \\
\textbf{Decomposition:} & 
\begin{lstlisting}[language=Scala,aboveskip=0cm,belowskip=-0.2cm,xleftmargin=0cm,style=dataexample]
Step 1: Find the event called "team meeting" that recurs weekly.
val s1 = theEvent(called("team meeting") and recurringWeekly)
Step 2: Find all events.
val s2 = findEvents0
Step 3: Filter events from list s2 to only include ones that intersect with event s1 that are not s1.
val s3 = s2.filter((x: Event) => x.interval.intersects(s1.interval) && x.id != s1.id)
Step 4: Decline events in s3.
val s4 = s3.map((x: Event) => respond(x, ResponseStatusType.declined)) \end{lstlisting} \\
\midrule
\textbf{Utterance 4:} & \textit{Find all the meetings this week where the duration is more than 2 hours and reschedule them to next week..} \\
\textbf{Decomposition:} & 
\begin{lstlisting}[language=Scala,aboveskip=0cm,belowskip=-0.2cm,xleftmargin=0cm,style=dataexample]
Step 1: Find all meetings happening this week
val s1 = findEvents(queryAt(`this`[Interval[Date]] and isWeek)) 
Step 2: Filter the list of events s1 to keep events with a duration greater than 2 hours
val s2 = s1.filter(x => x.duration > 2.hours) 
Step 3: Update each event in the list s2 to happen next week
val s3 = s2.map(x => modifyEvent(x, startsAt((next[Interval[Date]] and isWeek))))
\end{lstlisting} \\
        \bottomrule
    \end{tabular}
\caption{Examples of complex utterances in \ourdata{}. Each utterance is accompanied by decompositions consisting of a sequence of NL steps and associated program fragments, annotated by domain experts.
}
    \label{tab:data-examples-appendix}
\end{figure*}

\autoref{tab:data-examples-appendix} shows some example decompositions of complex utterances in \ourdata{}.

\section{Domain Library}
\label{appendix:domain_library}

The domain library defines the set of types and functions available in the domain.
Types model the domain objects such as \texttt{Person}, \texttt{Event},
whereas functions represent actions that can be taken by the agent,
including high-level APIs (\eg \texttt{createEvent}, \texttt{findEmails}),
low-level operations (\eg \texttt{min}, \texttt{+}),
predicate constructors (\eg \texttt{called}, \texttt{startsAt}),
etc.
The domain library is provided as Scala source code,
so that dataset users can statically validate generated code by compiling it with the domain library code.
Some builtin types (\eg String, Boolean), functions (\eg \texttt{map}, \texttt{Option.when}), 
control flow statements (\eg \texttt{if}) are not explicitly defined, but they can be used in the domain.

Natural language descriptions of entities or actions often omit some of their fields and focus on a subset of criteria that distinguish them from others.
In \ourdata{}, we represent these criteria as {\bf predicates}, 
which are lambda functions that take one or more arguments and return a \texttt{Boolean} value.
For example, a predicate that checks whether an event has the subject ``planning'' can be rewritten as
\texttt{called(\textrm{``}planning\textrm{''})},
where \texttt{called} is a predicate constructor defined in the domain library.
These predicate constructors simplify the annotations, avoiding spelling out the details of the field comparisons.
It also makes the program closer to natural language descriptions and potentially easier for LLMs to predict. 
To conjoin two predicates, the function \texttt{and} can be used. 

Further, the domain library provides a collection of extension methods and implicit conversions
which significantly simplify annotations for temporal expressions.
For example, the function \texttt{on} below can be used to 
combine a time expression and a date expression of different types.
\begin{lstlisting}[xleftmargin=0cm,linewidth=7.5cm]
extension [T](time: T)(using Conversion[T, LocalDateTime => Boolean]) {

  def on[U](date: U)(using Conversion[U, LocalDateTime => Boolean]): LocalDateTime => Boolean = ???

}
\end{lstlisting}
With this function and corresponding implicit conversions, 
``\textit{3pm on Monday}'' and ``\textit{morning on May 15}''
can be consistently annotated as \texttt{3.pm on Monday}
and \texttt{morning on (May /\& 15)}, respectively,
where \texttt{3.pm} returns \texttt{Time => Boolean},
\texttt{Monday} returns \texttt{DayOfWeek},
\texttt{morning} returns \texttt{Interval[Time]},
and \texttt{May /\& 15} returns \texttt{Date => Boolean}.
\section{Additional Method Details}

\subsection{Prompt Example}
\label{appendix:prompt_example}

\autoref{fig:prompt} shows a sample constructed prompt to generate a program fragment corresponding to the last generated step.  

\begin{figure*}
   \centering
\begin{lrbox}{\myv}\begin{minipage}{\textwidth}
\begin{lstlisting}[basicstyle=\fontsize{6}{7}\selectfont\ttfamily]
Step: Filter the list of events s2 to include only the events that are not the event s1
Program: val s3 = s2.filter(x => x.id != s1.id)
<EOS>

Step: Filter the list of events s3 to include only the events organized by me and which intersect with the event s1
Program: val s4 = s3.filter(x => x.organizerIs(me) && x.interval.intersects(s1.interval))
<EOS>

Step: Filter the list of persons s2 to include only people attending the event s1
Program: val s4 = s2.filter(x => s1.attendees.isAttending(x))
<EOS>

Step: Filter the list of persons s3 to include only the people attending the event s1
Program: val s5 = s3.filter(x => s1.attendees.isAttending(x))
<EOS>

Step: Filter the list of events s1 to only contain events with 1 other attendee
Program: val s = s1.filter(x => x.attendees.all.size == 2)
<EOS>

Step: Filter the list of days s1 to include only those days where there are at least 5 events
Program: val s = s1.filter(x => findEvents(queryAt(x /& (next[Interval[Date]] and isWeek))).size >= 5)
<EOS>

Step: Filter the list of events s5 to keep only those events that intersect with the interval s4
Program: val s6 = s5.filter(x => x.interval.intersects(s4))
<EOS>

Step: Filter the events in the list s1 to include only those where the other attendee has a job title of "PM"
Program: val s = s1.filter( x => getPersonFromAttendee(x.attendees.otherPeople.head).jobTitle == Some("PM"))
<EOS>

Step: Describe the number of events in the list s1
Program: val s2 = s1.size
<EOS>

Step: Update each event in the list s1 to last only for 30 minutes
Program: val s = s1.map(x => modifyEvent(x, lastsFor(30.minutes)))
<EOS>

Step: If the list s2 is empty then update the event s1 to end at 2:30 pm.
Program: val s3 = Option.when(s2.isEmpty){modifyEvent(s1, endsAt((2 :: 30).pm))}
<EOS>

Step: Filter the list of persons s1 to keep only those that don't have a one-on-one meeting with the user this week
Program: val s = s1.filter(x => findEvents(with_(x) and isOneOnOne and queryAt(`this`[Interval[Date]] and isWeek)).isEmpty)
<EOS>

Step: If there are more than 5 events in the list s1, decline the longest scheduled event
Program: val s = Option.when(s1.size > 5){respond(max(s1, (x => x.duration)), withResponse(ResponseStatusType.declined))}
<EOS>

Step: Change the event s1 to include Ben and remove Hao
Program: val s = modifyEvent(s1, with_("Ben") and not(with_("Hao")))
<EOS>

Step: Update the subject of each event in (list s1)[the list] to "1/1 with (corresponding person)"
Program: val s = s1.map(x => modifyEvent(x, called("1:1 with " + x.attendees.otherPeople.head.nameAndEmail.name.get)))
<EOS>

Step: Update each event in the list s2 to be called "project sync"
Program: val s = s2.map(x => modifyEvent(x, called("project sync")))
<EOS>

Step: Forward this email to the list of persons s2
Program: val s3 = forwardEmail(messageIdIs(theEmail(`this`[Email]).id) and messageWithTo(s2))
<EOS>

Step: Update each event in the list s3 to happen after the event s1
Program: val s5 = s4.map(x => modifyEvent(x, startsAt(after(s1.interval))))
<EOS>

Step: For each day in the list s2 create an event with the called "vacation time"
Program: val s = s2.map(x => createEvent(called("vacation time") and createAt(x /& (next[Interval[Date]] and isWeek))))
<EOS>

Step: Decline each meeting in the list s1
Program: val s = s1.map(x => respond(x, withResponse(ResponseStatusType.declined)))
<EOS>

Step: Check if the size of the list s1 is 5 or less
Program: val s = s1.size <= 5
<EOS>

Step: Update the event s1 to happen after 8am and before 5pm in the time zone s2.
Program: val s3 = modifyEvent(s1, startsAt(after(8.am) inZone s2) and endsAt(before(5.pm) inZone s2))
<EOS>

Step: Get the attendees of the event s1
Program: val s2 = s1.attendees.all
<EOS>

Step: Get the start time from event s1
Program: val s = s1.start.local.time
<EOS>

Step: Get the end time from event s1
Program: val s = s1.end.local.time
<EOS>



Utterance: Forward this email to all the attendees in the Standup event
Step 1: Find the event called Standup
Program 1: val s1 = theEvent(called("Standup"))
Step 2: Get the attendees of the event s1
Program 2: val s2 = s1.attendees.all
Step 3: Forward this email to the list of persons s2
Program 3: val s3 = forwardEmail(messageIdIs(theEmail(`this`[Email]).id) and messageWithTo(s2))
<EOS>

Utterance: If there is an "emergency review" meeting this week, then reschedule any events that happen 30 minutes before or after the meeting to next Friday.
Step 1: Find the event called "emergency review"
Program 1: val s1 = theEvent(called("emergency review"))
Step 2: Subtract 30 minutes from the start time of the event s1
Program 2: val s2 = s1.start - 30.minutes
Step 3: Add 30 minutes to the end time of the event s1
Program 3: val s3 = s1.end + 30.minutes
Step 4: Establish the time interval between the time instant s2 and the time instant s3
Program 4: val s4 = Interval[Instant](s2, s3)
Step 5: Find events that are scheduled for this week
Program 5: val s5 = findEvents(createAt(`this`[Interval[Date]] and isWeek))
Step 6: Filter the list of events s5 to keep only those events that intersect with the interval s4
Program 6: val s6 = s5.filter(x => x.interval.intersects(s4))
Step 7: Update each event in the list s6 to happen on the next Friday
Program 7: val s7 = s6.map(x => modifyEvent(x, createAt(next[Date] /& Friday)))
<EOS>

Utterance: Extend pizza party at 1 PM to end at 2.30 pm if extending it doesn't overlap with the next event.
Step 1: Find event called "pizza party"
Program 1: val s1 = theEvent(called("pizza party") and queryAt((1).pm))
Step 2: Find events starting between 1 pm and 2:30 pm that are not titled "pizza party"
Program 2: val s2 = findEvents(not(called("pizza party")) and queryAt(timeInterval(1.pm, (2::30).pm)))
Step 3: If the list s2 is empty then update the event s1 to end at 2:30 pm.
Program 3: val s3 = Option.when(s2.isEmpty){modifyEvent(s1, endsAt((2 :: 30).pm))}
<EOS>

Utterance: Create a preparation meeting this week with the attendees of the project sync who report to me or my manager.
Step 1: Find the event called "project sync"
Program 1: val s1 = theEvent(called("project sync"))
Step 2: Find my reports
Program 2: val s2 = thePerson(me).directReports
Step 3: Find my manager's reports
Program 3: val s3 = thePerson(me).manager.directReports
Step 4: Filter the list of persons s2 to include only people attending the event s1
Program 4: val s4 = s2.filter(x => s1.attendees.isAttending(x))
Step 5: Filter the list of persons s3 to include only the people attending the event s1
Program 5: val s5 = s3.filter(x => s1.attendees.isAttending(x))
Step 6: Create an event called preparation meeting this week with the list of persons s4 and the list of persons s5
Program 6: val s6 = createEvent(with_(s4) and with_(s5) and createAt(thisWeek) and called("preparation meeting"))
<EOS>

Utterance: Other than my manager, how many people are attending the project sync meeting tomorrow?
Step 1: Find the event called project sync tomorrow
Program 1: val s1 = theEvent(called("project sync") and queryAt(tomorrow))
Step 2: Describe the number of attendees of the event s1 excluding the manager
Program 2: val s2 = s1.attendees.all.size - 1
<EOS>

Utterance: Calculate how many meetings last week I had during my lunch hours of 12 noon to 1 PM
Step 1: Find events from last week between 12 PM and 1 PM
Program 1: val s1 = findEvents(queryAt(timeInterval(12.pm, 1.pm) on (last[Interval[Date]] and isWeek)))
Step 2: Describe the number of events in the list s1
Program 2: val s2 = s1.size
<EOS>

Utterance: Find all the meetings scheduled for next week that I created but that conflict with my doctor's appointment. Reschedule them to after the doctor's appointment.
Step 1: Find the event called "doctor's appointment"
Program 1: val s1 = theEvent(called("doctor's appointment"))
Step 2: Find all events happening next week
Program 2: val s2 = findEvents(queryAt(next[Interval[Date]] and isWeek))
Step 3: Filter the list of events s2 to include only the events that are not the event s1
Program 3: val s3 = s2.filter(x => x.id != s1.id)
Step 4: Filter the list of events s3 to include only the events organized by me and which intersect with the event s1
Program 4: val s4 = s3.filter(x => x.organizerIs(me) && x.interval.intersects(s1.interval))
Step 5: Update each event in the list s3 to happen after the event s1
Program 5: val s5 = s4.map(x => modifyEvent(x, startsAt(after(s1.interval))))
<EOS>

Utterance: Update the team meeting on Wednesday, so that its after 8 AM and before 5 PM for Jack
Step 1: Find the event called "team meeting" on Wednesday.
Program 1: val s1 = theEvent(called("team meeting") and queryAt(Wednesday))
Step 2: Find out what time zone Jack is in.
Program 2: val s2 = thePerson(named("Jack")).timeZone
Step 3: Update the event s1 to happen after 8am and before 5pm in the time zone s2.
Program 3: val s3 = modifyEvent(s1, startsAt(after(8.am) inZone s2) and endsAt(before(5.pm) inZone s2))
<EOS>

Utterance: How many 1/1 meetings in total I had in the last week?
Step 1: Find all one on one events from last week and return the size of that list
Program 1: val s1 = findEvents(queryAt(last[Interval[Date]] and isWeek) and isOneOnOne).size
<EOS>

Utterance: If I don't have an email about shiproom, then set up a 1:1 with Smith titled Discussion about Shiproom
Step 1: Check if there are no emails about shiproom
Program 1: val s1 = findEmails(messageTitleIs("shiproom")).isEmpty
Step 2: If s1 is true, set up a 1:1 with Smith titled "Discussion about Shiproom"
Program 2: val s2 = Option.when(s1){createEvent(isOneOnOne and with_("Smith") and called("Discussion about Shiproom"))}
<EOS>

Utterance: Adjust my schedule making sure there are no conflicts with the happy hour event today
Step 1: Find the event called "happy hour" today
Program 1: val s1 = theEvent(called("happy hour") and queryAt(today))
Step 2: Find all events happening today
Program 2: val s2 = findEvents(queryAt(today))
Step 3: Filter the list of events s2 to include only the events that intersect with the event s1
\end{lstlisting}
\end{minipage}\end{lrbox}
\resizebox{0.75\textwidth}{!}{\usebox\myv}
\caption{Example Prompt to generate program fragment for a generated step. The initial part of the prompt comprises of up to $M\leq 25$ examples similar to \textit{``Filter the list of events s2 to include only the events that intersect with the event s1''}. It is followed by $K=10$ decompositions of complex utterances. The generated output for the above prompt was \texttt{val s3 = s2.filter(x => x.interval.intersects(s1.interval))} }
\label{fig:prompt}
\end{figure*}

\subsection{LLM APIs}
We use OPEN-AI's APIs, as per their terms of use \url{https://openai.com/policies/terms-of-use}
\section{Additional Details on Experiments}
\label{appendix:experiments}

\subsection*{Experiments with Elementary Utterances}
\newremove{In this section, we provide additional details for NL to program experiments on elementary utterances. As mentioned previousxly, for this experiment, we split the elementary utterances data, consisting of 841 utterances, into train, dev and test splits in the ratio 70:15:15.}\new{To contextualize model performance on complex
utterances, we conclude by analyzing how the NL-to-program semantic parser fares on elementary utterances in the DeCU dataset. We split the elementary utterances data, consisting of 841 utterances, into train, dev and test splits in the ratio 70:15:15.} We manually tried a few tweaks (about 5 variations were tried) to the prompt structure and varied parameters such as the number of exemplars in the prompt, and picked the setup that resulted in the highest exact match accuracy on the dev split. 
We report an exact match against the gold parse as has been used in the past work as well \cite{DBLP:journals/corr/abs-2206-10668}. Additionally, we note that there might be certain small deviations such as extra surrounding braces that do not invalidate the generated program, and exact match as a binary metric would penalize such deviations. So we additionally report a character-based edit distance measure \cite{wang-etal-2016-character} that might provide more fine-grained insights compared to binary exact match.

\begin{table}[t]
    \centering
    \footnotesize

\begin{tabular}{lcc}
\toprule
\textbf{Method} & \textbf{Exact Match}$\uparrow$ & \textbf{Char Edit}$\downarrow$ \\
 \midrule 
LLM parser & 0.60 & 0.10 \\
\multicolumn{2}{l}{\quad {\bf Ablations}:} \\
Using text-davinci-001 & 0.42 & 0.23 \\
Random exemplars & 0.21 & 0.41 \\
Max 3 examples & 0.60 & 0.13 \\
Limit to 20\% train data & 0.33 & 0.26 \\
\bottomrule
    \end{tabular}
    \caption{Evaluation results on parsing elementary utterances in the test set.
    }
    \label{tab:calendar_results}
\end{table}

\noindent \textbf{Results:}
Results on the parsing task for the test split are shown in \autoref{tab:calendar_results}. The parser, referred to as LLM Parser in the Table, gets $0.60$ exact match and $0.10$ character edit distance. 
We also consider a few variations of the parser: \\
(1) Using \emph{text-davinci-001} instead of \emph{text-davinci-003}: Leads to a large reduction in the automated metrics.   \\
(2) Using random $M$ examples in the prompt instead of dynamically selected ones based on similarity to test command: Performs the worst among all the variations, demonstrating that relevant example selection is very important. \\
(3) Reducing the number of in-context examples from $M=20$ to $M=3$: The metrics show little to no change, suggesting that just a couple of relevant examples are usually enough for the parser.  \\
(4) Reduce train data size to around 20\% of the original size: This leads to a reduction in accuracy values, as in-context learning relies on having similar relevant examples in the prompt. 

As mentioned earlier, a limitation of the exact match metric is that it is not perfect, since it fails to capture extra redundant braces, or small variations in string argument values, such as the meeting name being ``coffee meeting'' instead of ``coffee''.
We observe that some of the incorrect predictions are due to a few functions in gold program not being present in the training data (our splits were random). 
Using API documentation and containing the outputs of the parser \cite{DBLP:conf/emnlp/ShinLTCRPPKED21, DBLP:journals/corr/abs-2206-10668} to only use allowed functions could be leveraged to fix such errors, though we leave it for future work, as our primary goal is to study complex utterance parsing in light of training data for simpler elementary utterances.

\subsection*{Additional Example Outputs}

\begin{figure*}[t]
    \centering
    \footnotesize
    \begin{tabular}{>{\centering\arraybackslash}m{2cm} m{12cm}}
    \toprule

    \textbf{Utterance 1:}  
    & \textit{Make me a meeting at 3 pm mountain time and refuse any meetings that happen at the same time.} \\ 
    
    \textbf{\method{}}
    & \begin{lstlisting}[language=Scala,aboveskip=0cm,belowskip=-0.2cm,xleftmargin=0cm,style=dataexample]
Step 1: Find events at 3 pm mountain time
val s1 = findEvents(queryAt(3.pm inZone TimeZone("MST"))) 
Step 2: Refuse any events in the list s1
val s2 = s1.map(x => respond(x, withResponse(ResponseStatusType.declined)))
Step 3: Create a meeting at 3 pm mountain time
val s3 = createEvent(createAt(3.pm inZone TimeZone("MST"))) 
\end{lstlisting} \\

    \textbf{\nostepmethod{}}
    & \begin{lstlisting}[language=Scala,aboveskip=0cm,belowskip=-0.2cm,xleftmargin=0cm,style=dataexample]
val s1 = createEvent(createAt(3.pm inZone TimeZone("MST"))) 
val s2 = findEvents(queryAt(3.pm inZone TimeZone("MST"))) 
val s3 = s2.map(x => refuseEvent(x))
\end{lstlisting}  \\
    
    \textbf{Comment} 
    & We observe that hallucination of functions is much more prominent in the baseline \nostepmethod{},
    as can be seen in the use of the non-existing function \texttt{refuseEvent} \\

    \midrule
    
    \textbf{Utterance 2:} & \textit{Cross all my vacation events off my calendar, I need to juggle those around.} \\
    
    \textbf{\method{}} 
    & \begin{lstlisting}[language=Scala,aboveskip=0cm,belowskip=-0.2cm,xleftmargin=0cm,style=dataexample]
Step 1: Find all events called "vacation"
val s1 = findEvents(called("vacation")) 
Step 2: Delete all events in the list s1
val s2 = s1.map(x => deleteEvent(x))
\end{lstlisting} \\
 
    \textbf{\nostepmethod{}}
    & \begin{lstlisting}[language=Scala,aboveskip=0cm,belowskip=-0.2cm,xleftmargin=0cm,style=dataexample]
val s = theEvent(called("vacation")).map(x => deleteEvent(x)) 
\end{lstlisting} \\

    \textbf{Comment} 
    & Prediction from \nostepmethod{} incorrectly applies a \texttt{map} operation to a single event. We note that predictions from \method{} show fewer syntax errors compared to \nostepmethod{}. \\

    \midrule
    
    \textbf{Utterance 3:}  
    & \textit{If Issac cannot make it to the LT-meeting, schedule some 1:1 time with him tonight.} \\ 
    
    \textbf{\method{}}
    & \begin{lstlisting}[language=Scala,aboveskip=0cm,belowskip=-0.2cm,xleftmargin=0cm,style=dataexample]
Step 1: Find the event called "LT-meeting"
val s1 = theEvent(called("LT-meeting")) 
Step 2: Check if Issac is attending the event s1
val s2 = s1.attendees.isAttending(thePerson("Issac")) 
Step 3: If s2 is false, create a 1:1 event with Issac tonight
val s3 = Option.when(!s2){createEvent(isOneOnOne and with_("Isaac") and createAt(tonight))}
\end{lstlisting} \\

    \textbf{\nostepmethod{}}
    & \begin{lstlisting}[language=Scala,aboveskip=0cm,belowskip=-0.2cm,xleftmargin=0cm,style=dataexample]
val s1 = theEvent(called("LT-meeting")) 
val s2 = thePerson(named("Issac")) 
val s3 = s1.attendees.isAttending(s2) 
val s4 = if (s3) { 
createEvent(with_(s2) and createAt(tonight) and called("1:1 time")) 
} else { 
null 
} 
\end{lstlisting}  \\
    
    \textbf{Comment} 
    & \nostepmethod{} output has a logical flaw --- it creates a meeting when Issac is attending the `LT-meeting' -- and was judged as incorrect. We observe that outputs from \method{} are less likely to demonstrate such logical flaws. \\

    \midrule
    
    \textbf{Utterance 4:} & \textit{I need you to delete the meeting I have at 3 pm and replace it with "Core" with Jimmy.} \\
    
    \textbf{\method{}} 
    & \begin{lstlisting}[language=Scala,aboveskip=0cm,belowskip=-0.2cm,xleftmargin=0cm,style=dataexample]
Step 1: Find the event at 3 pm
val s1 = theEvent(queryAt(3.pm)) 
Step 2: Delete the event s1
val s2 = deleteEvent(s1) 
Step 3: Create a new event called "Core with Jimmy" at 3 pm
val s3 = createEvent(called("Core with Jimmy") and createAt(3.pm))
\end{lstlisting} \\
    
    \textbf{\nostepmethod{}}
    & \begin{lstlisting}[language=Scala,aboveskip=0cm,belowskip=-0.2cm,xleftmargin=0cm,style=dataexample]
val s1 = deleteEvent(theEvent(queryAt(3.pm))) 
val s2 = createEvent(called("Core") and with_("Jimmy"))
\end{lstlisting} \\
    
    \textbf{Comment} 
    & Prediction from \nostepmethod{} fails to create the meeting at the correct time, 
    and was judged as incorrect in human evaluations. \\

    \midrule
    
    \textbf{Utterance 5:}  
    & \textit{Rename all the 1/1 meetings I have with a PM this month to be called project sync.} \\ 
    
    \textbf{\method{}}
    & \begin{lstlisting}[language=Scala,aboveskip=0cm,belowskip=-0.2cm,xleftmargin=0cm,style=dataexample]
Step 1: Find all 1/1 meetings this month
val s1 = findEvents(queryAt(`this`[Interval[Date]] and isMonth) and isOneOnOne) 
Step 2: Filter the list of events s1 to include only the events with a PM
val s2 = s1.filter( x => getPersonFromAttendee(x.attendees.otherPeople.head).jobTitle == Some("PM")) 
Step 3: Update each event in the list s2 to be called "project sync"
val s3 = s2.map(x => modifyEvent(x, called("project sync"))) 
\end{lstlisting} \\

    \textbf{\nostepmethod{}}
    & \begin{lstlisting}[language=Scala,aboveskip=0cm,belowskip=-0.2cm,xleftmargin=0cm,style=dataexample]
val s1 = findEvents(queryAt(`this`[Interval[Date]] and isMonth) and isOneOnOne) 
val s2 = s1.filter(x => x.attendees.isAttending(thePerson(PM))) 
val s3 = s2.map(x => modifyEvent(x, called("project sync")))
\end{lstlisting}  \\
    
    \textbf{Comment} 
    & \nostepmethod{} output was not valid as per library since `thePerson(PM)`  is incorrect usage while \method{} output was judged as correct.  \\

    \bottomrule

    \end{tabular}
    
    \caption{Outputs from \method{} and \nostepmethod{} on a few complex utterances in \ourdata{}.}
    \label{tab:qual-examples-appendix}
\end{figure*}

\begin{figure*}[t]
    \centering
    \footnotesize
    \begin{tabular}{>{\centering\arraybackslash}m{2cm} m{12cm}}
    \toprule

    \textbf{Utterance:}  
    & \textit{Make me a meeting at 3 pm mountain time and refuse any meetings that happen at the same time.} \\ 
    
    \textbf{\method{}}
    & \begin{lstlisting}[language=Scala,aboveskip=0cm,belowskip=-0.2cm,xleftmargin=0cm,style=dataexample]
Step 1: Find events at 3 pm mountain time
val s1 = findEvents(queryAt(3.pm inZone TimeZone("MST"))) 
Step 2: Refuse any events in the list s1
val s2 = s1.map(x => respond(x, withResponse(ResponseStatusType.declined)))
Step 3: Create a meeting at 3 pm mountain time
val s3 = createEvent(createAt(3.pm inZone TimeZone("MST"))) 
\end{lstlisting} \\

    \textbf{\nostepmethod{}}
    & \begin{lstlisting}[language=Scala,aboveskip=0cm,belowskip=-0.2cm,xleftmargin=0cm,style=dataexample]
val s1 = createEvent(createAt(3.pm inZone TimeZone("MST"))) 
val s2 = findEvents(queryAt(3.pm inZone TimeZone("MST"))) 
val s3 = s2.map(x => refuseEvent(x))
\end{lstlisting}  \\

    \textbf{\cotbaseline{}}
    & \begin{lstlisting}[language=Scala,aboveskip=0cm,belowskip=-0.2cm,xleftmargin=0cm,style=dataexample]
Step 1: Create a meeting at 3 pm mountain time
Step 2: Refuse any meetings that happen at the same time"}
val s1 = createEvent(createAt(3.pm inZone TimeZone(\"MST\")))
val s2 = findEvents(queryAt(3.pm inZone TimeZone(\"MST\"))).map(x => refuseEvent(x))
\end{lstlisting}  \\

    \bottomrule

    \end{tabular}
    
    \caption{Outputs from \method{}, \nostepmethod{} and \cotbaseline{} on a sample complex utterance in \ourdata{}.}
    \label{tab:qual-examples-appendix-all}
\end{figure*}

\begin{figure*}[t]
    \centering
    \footnotesize
    \begin{tabular}{>{\centering\arraybackslash}m{2cm} m{12cm}}
    \toprule
    
    \textbf{Utterance 1:} & \textit{Split my project sync call next week into two events with the same attendees, but in the second one, invite David Smith also.} \\
    
    \textbf{\method{}} 
    & \begin{lstlisting}[language=Scala,aboveskip=0cm,belowskip=-0.2cm,xleftmargin=0cm,style=dataexample]
Step 1: Find the event called "project sync" next week
Plan 1: val s1 = theEvent(called("project sync") and queryAt(next[Interval[Date]] and isWeek)) 
Step 2: Get the attendees of the event s1
Plan 2: val s2 = s1.attendees.all 
Step 3: Create a new event with the same attendees as the event s1
Plan 3: val s3 = createEvent(with_(s2) and createAt(s1.interval)) 
Step 4: Add David Smith to the list of attendees s2
Plan 4: val s4 = s2 :+ thePerson(named("David Smith")) 
Step 5: Create a new event with the list of attendees s4
Plan 5: val s5 = createEvent(with_(s4) and createAt(s1.interval))
\end{lstlisting} \\
 
    \textbf{\nostepmethod{}}
    & \begin{lstlisting}[language=Scala,aboveskip=0cm,belowskip=-0.2cm,xleftmargin=0cm,style=dataexample]
val s1 = theEvent(called("project sync") and queryAt(next[Interval[Date]] and isWeek)) 
val s2 = createEvent(with_(s1.attendees.all) and createAt(s1.interval.start) and called("project sync 1")) 
val s3 = createEvent(with_(s1.attendees.all) and with_("David Smith") and createAt(s1.interval.end) and called("project sync 2"))
\end{lstlisting} \\

\textbf{Experts}
    & \begin{lstlisting}[language=Scala,aboveskip=0cm,belowskip=-0.2cm,xleftmargin=0cm,style=dataexample]
Step 1: Find the "project sync" event next week
val s1 = theEvent(called("project sync") and queryAt(next[Interval[Date]] and isWeek)) 
Step 2: Create an event with the same attendees from event s1 and include David Smith
val s2 = createEvent(with_(s1.attendees) and with_("David Smith"))
\end{lstlisting} \\

    \textbf{Comment} 
    & Output program from \method{} as well as the output from \nostepmethod{} doesn't delete the original meeting and creates two new meetings, which does not reflect the user intention. \\

    \midrule

    \textbf{Utterance 2:} & \textit{I'd like to divide the meeting at noon in two, so split off the last fifteen minutes as a separate event.} \\
    
    \textbf{\method{}} 
    & \begin{lstlisting}[language=Scala,aboveskip=0cm,belowskip=-0.2cm,xleftmargin=0cm,style=dataexample]
Step 1: Find the event at noon
val s1 = theEvent(createAt(noon)) 
Step 2: Subtract 15 minutes from the end time of the event s1
val s2 = s1.end - 15.minutes 
Step 3: Update the event s1 to end at the time instant s2
val s3 = modifyEvent(s1, endsAt(s2)) 
Step 4: Create a new event starting at the time instant s2
val s4 = createEvent(startsAt(s2))
\end{lstlisting} \\
 
    \textbf{\nostepmethod{}}
    & \begin{lstlisting}[language=Scala,aboveskip=0cm,belowskip=-0.2cm,xleftmargin=0cm,style=dataexample]
val s1 = theEvent(createAt(noon)) 
val s2 = modifyEvent(s1, endsAt(s1.start - 15.minutes)) 
val s3 = createEvent(startsAt(s1.start - 15.minutes) and lastsFor(15.minutes))
\end{lstlisting} \\

\textbf{Experts}
    & \begin{lstlisting}[language=Scala,aboveskip=0cm,belowskip=-0.2cm,xleftmargin=0cm,style=dataexample]
Step 1: Find my event at noon
val s1 = theEvent(queryAt(noon)) 
Step 2: Reduce the duration of event s1 by 15 minutes
val s2 = modifyEvent(s1, lastsFor(s1.duration - 15.minutes)) 
Step 3: Get the end time of event s2
val s3 = s2.end.local.time 
Step 4: Create an event that starts at the end of event s3 that lasts for 15 minutes
val s4 = createEvent(createAt(s3) and lastsFor(15.minutes))
\end{lstlisting} \\

    \textbf{Comment} 
    & Output program from \method{} was judged as incorrect since it doesn't specify the duration of the second event. \\
    
    \bottomrule

    \end{tabular}
    
    \caption{Sample predictions from \method{} that were judged as incorrect in human evaluations.}
    \label{tab:qual-examples-appendix-errors}
\end{figure*}

\autoref{tab:qual-examples-appendix} shows predictions from \method{} and \nostepmethod{} on a few sample inputs.
\autoref{tab:qual-examples-appendix-all} shows predictions from \method{} and the two main baselines \nostepmethod{} and \cotbaseline{} on a sample input.
\autoref{tab:qual-examples-appendix-errors} shows a few cases where outputs from \method{} were incorrect.

\newremove{\subsection*{Details about Automated Evaluation of NL Decomposition} 
To gauge the efficacy of the automated metric, we compared it to the manual evaluation performed by one of the authors on a third of the data (predictions and expert annotations).  On this third, expert annotations and \method{} predictions got rated as 97.5\% and 85.0\% correct, respectively, as reported in \autoref{tab:auto_results}. The automatic labels agree with the manual ones 90\% of the time.
Moreover, Cohen's Kappa coefficient between the two sets of ratings (manual annotations and automatic ones) is 0.28, indicating that automated labels seem to have a fair correlation with manual ratings, even though automated evaluation does seem a bit biased towards the model-generated predictions. The prompt used for evaluation will be released along with the code, as it provides scalable and reproducible evaluation. }

\begin{table}[!t]
    \centering
    \footnotesize
    \begin{tabular}{l|c|c}
        \toprule
        & \textsc{Experts} & \method{} \\
        \midrule
        Grammar                 & 98.3 & 98.6 \\
        Factuality              & 99.4 & 99.4 \\
        Hallucination           & 99.4 & 99.4 \\
        Redundancy              & 99.7 & 99.7 \\
        Repetition              & 100 & 100 \\
        Missing Information     & 99.7 & 97.8 \\
        Coherency               & 98.8 & 99.7 \\
        Commonsense             & 100 & 100 \\
        Arithmetic              & 100 & 100 \\
        \bottomrule
    \end{tabular}
    \caption{Analysis of the NL steps written by domain experts 
    and those predicted by \method{}.
    For the overall evaluation, the score is the percentage of examples judged as
    \textit{sufficient and correct} to complete the user request.
    For the step-by-step evaluation done by crowd workers, the score of a dimension is the percentage of 
    steps judged as not containing related issues. 
    }
    \label{tab:mturk_eval}
\end{table}
\subsection*{Step-by-Step Evaluation of NL Decomposition:}
\new{
Following ROSCOE \cite{golovneva-etal-2023-roscoe},\footnote{
We adopted ROSCOE as much as possible, though we received some feedback 
from crowdworkers that some dimensions (such as factuality, hallucination, and coherency) were difficult to judge for our evaluation task.}
we evaluate the quality of individual steps on 9 dimensions: grammar, factuality, hallucination, redundancy, repetition, missing information, coherency, commonsense, and arithmetic.
We recruited annotators from  Amazon Mechanical Turk to provide binary classification ratings (\textit{yes} or \textit{no}) for each step in the decomposition on all 9 dimensions.
For each question, 3 judgments are collected and the majority-voted answer is used as the final judgment.
For quality control, we restrict to annotators located in the United States or Canada
who have an approval rate higher than $85\%$
and have successfully solved a qualification task where we match their answers on the same set of questions against answers manually annotated by one of the authors. We pay 0.25 USD per example per question.
The step-by-step evaluation results shown in the bottom block of \autoref{tab:mturk_eval} 
suggest that the quality of individual steps is very high.
\method{}-predicted steps are rated similar to expert steps on almost all dimensions,
except on the ``missing information'' dimension, where the gap is noticeable.
Note that it is possible for all steps in a program to be judged individually correct, but fail to complete the user request. %
}

\subsection*{Train and Test Split of Complex Utterances } 

As noted previously, we use $K=10$ complex utterance examples from \ourdata{} as train split. The examples were chosen randomly. We investigate the impact of random seed, by repeating the experiments for 3 random seeds (i.e. each seed leads to a different set of $10$ examples in the train split while the remaining are in the test split), and report the average automated scores in \autoref{tab:appendix:reruns}.

\begin{table}[t]
    \centering
    \footnotesize
    \begin{tabular}{l@{\hskip 0.01in}cc}
        \toprule
        \textbf{System} & \textbf{Correct}$\uparrow$ & \textbf{CER}$\downarrow$ \\ %
        \midrule 
         \nostepmethod{} & $0.31$ & $0.46$ \\ %
         \method{} & $\bf0.41$ & $\bf0.41$ \\ %
        \bottomrule
    \end{tabular}
    \caption{Automated Metrics averaged over 3 runs.
    }
    \label{tab:appendix:reruns}
\end{table}

\new{
\subsection*{Additional Details on Correct Reference-less Metric using GPT4} 

\textbf{Correlation of model-based metrics with human labels:} 
On a per-example basis, we observe that automatic labels and human evaluation labels agree on correct/incorrect annotation 96.5\% of the time when compared on the subset of well-formed outputs from the above-mentioned methods. Moreover, considering human-provided labels as ground truth and \textbf{Correct} labels as predictions in a binary classification setup, we observe F1 scores of 98\% and 84\% for ``correct" and ``incorrect" classes respectively. Thus, \textbf{Correct} seems to be a good proxy for a human label to judge the correctness of a generated program. 

\textbf{Prompt:} Prompt used in the reference-less evaluation is shown in Figure \ref{fig:gptevalprompt}.

}

\begin{figure*}
   \centering
\begin{lrbox}{\myv}\begin{minipage}{\textwidth}
\begin{lstlisting}[basicstyle=\fontsize{8}{10}\selectfont\ttfamily]

Task is to identify whether a program, written in a Scala-like representation, is sufficient and correct to complete the user request. 
The program can be considered to be an incorrect solution for various reasons such as if it addresses only a part of user request, or does more than what user asked for, or uses incorrect values or variable names, etc. 
The program can be considered to be correct if it reflects the user intent, and nothing more, nothing less.


# Utterance: Find all the meetings scheduled for next week that I created but that conflict with my doctor's appointment. Reschedule them to after the doctor's appointment.
# Predicted program
val s1 = findEvents(called("doctor")).get(0)
val s2 = findEvents(queryAt(next[Interval[Date]] and isWeek))
val s3 = s2.filter(x => x.organizerIs(me) && x.interval.intersects(s1.interval) && x.id != s1.id)
val s4 = s3.map(x => modifyEvent(x, startsAt(after(s1.end))))
# Judge whether predicted program is correct or not
## Description of predicted program:
s1: Find events about "doctor"
s2: Find events this week
s3: Filter the events in s2 to keep only those that are organized by me and intersect with s1 and are not s1
s4: Modify the events in list s3 to happen after event s1
## Utterance: Find all the meetings scheduled for next week that I created but that conflict with my doctor's appointment. Reschedule them to after the doctor's appointment.
## Reflection on predicted program:
In this example, the predicted program correctly reflects the user intent in the utterance.
## Predicted program is: Correct
<EOS>


# Utterance: If I don't have an email about shiproom, then set up a 1:1 with Smith titled Discussion about Shiproom
# Predicted program
val s1 = findEmails(messageTitleIs("shiproom"))
val s2 = e.isEmpty
# Judge whether predicted program is correct or not
## Description of predicted program:
s1: Find the emails about "shiproom"
s2: Check is the list s1 is empty or not
## Utterance: If I don't have an email about shiproom, then set up a 1:1 with Smith titled Discussion about Shiproom
## Reflection on predicted program:
In this example, the predicted program does not create 1:1 meeting as per the user intent from the utterance. Thus, it is not sufficient.
## Predicted program is: Incorrect
<EOS>


# Utterance: Update the team meeting on Wednesday, so that its after 8 AM and before 5 PM for Jack
# Predicted program
val s1 = theEvent(called(\"team meeting\") and queryAt(Wednesday))
val s2 = thePerson(named(\"Jack\")).timeZone
val s3 = modifyEvent(s1, startsAt(after(8.am) inZone s2) and endsAt(before(5.pm) inZone s2))
# Judge whether predicted program is correct or not
## Description of predicted program:
s1: Find the event called "team meeting" on Wednesday.
s2: Find out what time zone Jack is in
s3: Update the event s1 to happen after 8am and before 5pm in the time zone s2.
## Utterance: Update the team meeting on Wednesday, so that its after 8 AM and before 5 PM for Jack
## Reflection on predicted program:
In this example, the predicted program correctly reflects the user intent in the utterance.
## Predicted program is: Correct
<EOS>


# Utterance: Create a preparation meeting this week with the attendees of the project sync who report to me or my manager.
# Predicted program
val s1 = theEvent(called("project sync"))
val s2 = thePerson(me).directReports
val s4 = s2.filter(x => s1.attendees.isAttending(x))
val s6 = createEvent(with_(s4) and createAt(thisWeek) and called("preparation meeting"))
# Judge whether predicted program is correct or not
## Description of predicted program:
s1: Find the event called "project sync"
s2: Find my reports
s4: Filter the list of persons s2 to include only people attending the event s1
s6: Create an event called preparation meeting this week with the list of persons s4
## Utterance: Create a preparation meeting this week with the attendees of the project sync who report to me or my manager.
## Reflection on predicted program:
In this example, the predicted program does not consider the people who report to user's manager.  Thus, it is not sufficient.
## Predicted program is: Incorrect
<EOS>

# Utterance: <test-utterance>
# Predicted Program: <predicted program>
# Judge whether predicted program is correct or not

\end{lstlisting}
\end{minipage}\end{lrbox}
\resizebox{0.75\textwidth}{!}{\usebox\myv}
\caption{Prompt used to compute Correct metric using GPT4 .}
\label{fig:gptevalprompt}
\end{figure*}

\begin{figure*}[]
    \centering
    \includegraphics[width=.98\textwidth,clip,trim=0in 6in 3in 0in]{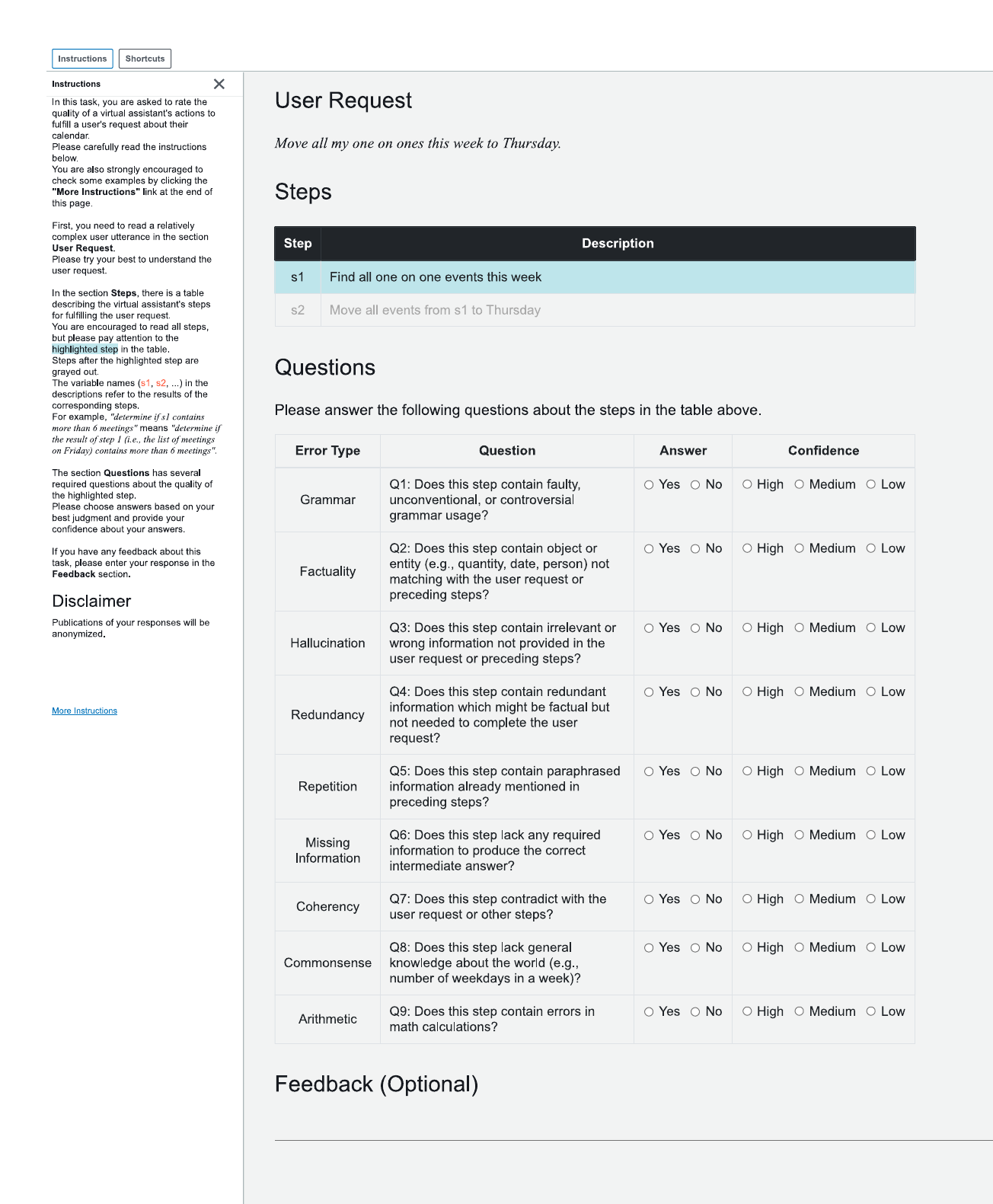}
    \caption{The annotation interface for the step-by-step evaluation on the NL steps.}
    \label{fig:mturk_eval_steps}
\end{figure*}

\end{document}